\documentclass[11pt]{article}
\usepackage{amsgen,amsmath,amstext,amsbsy,amsopn,amssymb,mathabx,mathrsfs,amsthm,bm,bbm,romannum,outlines,enumitem,natbib,xr,multirow}
\usepackage{caption}
\usepackage[dvips]{graphicx}
\usepackage{subcaption}
\usepackage{diagbox}
\usepackage[ruled,vlined,linesnumbered,resetcount]{algorithm2e}

\usepackage{hyperref}
\hypersetup{
	colorlinks=true,
	linkcolor=red,
	urlcolor=blue,
	citecolor=blue
}
\usepackage[depth=2]{bookmark}

\usepackage[margin=1in]{geometry}

\usepackage[utf8]{inputenc}

\usepackage{booktabs}
\usepackage{longtable}
\usepackage{array}
\usepackage{multirow}
\usepackage[table]{xcolor}
\usepackage{wrapfig}
\usepackage{float}
\usepackage{colortbl}
\usepackage{pdflscape}
\usepackage{tabu}
\usepackage{threeparttable}
\usepackage{threeparttablex}
\usepackage[normalem]{ulem}
\usepackage{makecell}

\newtheorem{thm}{Theorem}
\newtheorem{lem}[thm]{Lemma}
\newtheorem{prop}[thm]{Proposition}
\newtheorem{coro}[thm]{Corollary}

\theoremstyle{definition}

\newtheorem{asm}{Assumption}

\newtheorem{rmk}{Remark}

\theoremstyle{remark}

\newcommand{\bx}{\bm{x}}

\newcommand{\Ib}{\mathbf{I}}

\newcommand{\bX}{\bm{X}}

\newcommand{\bbeta}{\bm{\beta}}

\newcommand{\bmu}{\bm{\mu}}

\newcommand{\bbR}{\mathbb{R}}
\newcommand{\bbE}{\mathbb{E}}

\newcommand{\bbP}{\mathbb{P}}

\newcommand{\bbQ}{\mathbb{Q}}

\newcommand{\cA}{\mathcal{A}}

\newcommand{\cC}{\mathcal{C}}
\newcommand{\cD}{\mathcal{D}}

\newcommand{\cF}{\mathcal{F}}

\newcommand{\cL}{\mathcal{L}}

\newcommand{\cN}{\mathcal{N}}
\newcommand{\cO}{\mathcal{O}}
\newcommand{\cP}{\mathcal{P}}

\newcommand{\cR}{\mathcal{R}}

\newcommand{\cV}{\mathcal{V}}

\newcommand{\cX}{\mathcal{X}}
\newcommand{\cY}{\mathcal{Y}}

\newcommand{\sP}{\mathscr{P}}

\newcommand{\argmin}{\mathop{\mathrm{argmin}}}

\newcommand{\VaR}{\mathrm{VaR}}
\newcommand{\CVaR}{\mathrm{CVaR}}

\newcommand{\bsign}{\textbf{sign}}

\newcommand{\mix}{\mathrm{mix}}		

\newcommand{\IPWE}{\mathrm{IPWE}}
\newcommand{\CTE}{\mathrm{CTE}}

\newcommand{\rand}{\mathrm{rand}}
\newcommand{\test}{\mathrm{test}}

\newcommand{\EL}{\mathrm{EL}}		
\newcommand{\KL}{\mathrm{KL}}		
\newcommand{\CR}{\mathrm{CR}}		

\newcommand{\bzero}{{\mathbf{0}}}	
\newcommand{\bbone}{{\mathbbm{1}}}	

\newcommand{\rd}{\mathrm{d}}		

\newcommand\indep{\protect\mathpalette{\protect\independenT}{\perp}}
\def\independenT#1#2{\mathrel{\rlap{$#1#2$}\mkern2mu{#1#2}}}	

\newcommand{\Bern}{\textbf{Bernoulli}}

\begin{title}
	{ Learning Optimal Distributionally Robust Individualized Treatment Rules }
\end{title}
\author{
	Weibin Mo, Zhengling Qi and Yufeng Liu\thanks{Weibin Mo and Zhengling Qi are co-first authors for the paper. Weibin Mo is a Ph.D. student, Department of Statistics and Operations Research, University of North Carolina at Chapel Hill, Chapel Hill, NC 27599, USA. E-mail: \href{mailto:harrymok@email.unc.edu}{harrymok@email.unc.edu}. Zhengling Qi is Assistant Professor, Department of Decision Sciences, George Washington University, Washington, D.C. 20052, USA. E-mail: \href{mailto:qizhengling@gwu.edu}{qizhengling@gwu.edu}. Yufeng Liu is Professor, Department of Statistics and Operations Research, Department of Genetics, Department of Biostatistics, Carolina Center for Genome Science, Lineberger Comprehensive Cancer Center, University of North Carolina at Chapel Hill, NC 27599, USA. E-mail: \href{mailto:yfliu@email.unc.edu}{yfliu@email.unc.edu}.}
}
\date{}
\begin{document}
	\pagenumbering{arabic}
	\maketitle
	
	
	\begin{abstract}
		Recent development in the data-driven decision science has seen great advances in individualized decision making. Given data with individual covariates, treatment assignments and outcomes, policy makers best individualized treatment rule (ITR) that maximizes the expected outcome, known as the value function. Many existing methods assume that the training and testing distributions are the same. However, the estimated optimal ITR may have poor generalizability when the training and testing distributions are not identical. In this paper, we consider the problem of finding an optimal ITR from a restricted ITR class where there is some unknown covariate changes between the training and testing distributions. We propose a novel distributionally robust ITR (DR-ITR) framework that maximizes the worst-case value function across the values under a set of underlying distributions that are ``close'' to the training distribution. The resulting DR-ITR can guarantee the performance among all such distributions reasonably well. We further propose a calibrating procedure that tunes the DR-ITR adaptively to a small amount of calibration data from a target population. In this way, the calibrated DR-ITR can be shown to enjoy better generalizability than the standard ITR based on our numerical studies.
	\end{abstract}
	\newpage
	\setcounter{page}{1}
	\section{Introduction}\label{sec:intro}
	Data-driven individualized decision making problems are commonly seen in practice and have been studied intensively in the literature. In disease management, the physician may decide whether to introduce or switch a therapy for a patient based on his/her characteristics in order to achieve a better clinical outcome \citep{bertsimas2017personalized}. In public policy making, a policy that allocates the resource based on the characteristics of the targets can improve the overall resource allocation efficiency \citep{kube2019allocating}. In a context-based recommender system, the use of the contextual information such as time, location and social connection can increase the effectiveness of the recommendation process \citep{aggarwal2016recommender}. One common goal of these problems is to find the optimal \textit{individualized treatment rule (ITR)} mapping from the individual characteristics or contextual information to the treatment assignment, that maximizes the expected outcome, known as the \textit{value function} \citep{manski2004statistical,qian2011performance}. 
	
	One approach for estimating an optimal ITR is to first estimate the conditional mean outcome, known as the \textit{$ Q $-function}, given the individual characteristics and the treatment assignment, and then induce the ITR that prescribes the treatment by maximizing the estimated $ Q $-function \citep{qian2011performance}. In the binary treatment case, such an approach can be reformulated as estimating the \textit{conditional treatment effect (CTE)} as the difference of the conditional mean outcomes under two candidate treatments \citep{zhao2017selective,chen2017general,qi2020multi}. Another approach is to directly estimate the value function using the \textit{inverse-probability weighted estimator (IPWE)}, and then search for the ITR that maximizes the corresponding  value function \citep{zhao2012estimating,kitagawa2018should,liu2018augmented,zhang2019multicategory}. Since there are potential model misspecification issues of these approaches, the \textit{augmented IPWE (AIPWE)} of the value function combines the estimates of the $ Q $-function and the treatment propensity score. AIPWE is \textit{doubly robust} in the sense that the consistency of the value function estimate is guaranteed as long as either the $ Q $-function model or the propensity score model is correctly specified	\citep{dudik2011doubly,zhang2012robust,athey2017efficient,zhao2019efficient}. While the doubly robust property can protect against the violation of the model assumptions, one key assumption behind is that the training and testing distributions should be identical.
	
	When the training and testing distributions are different, an estimated optimal ITR may not generalize well on the testing data \citep{zhao2019robustifying}. Similar phenomenon for causal inference in randomized controlled trials (RCTs) has also been pointed out by \citet{muller2014randomised,gatsonis2017methods}. Specifically, due to the inclusion and exclusion criteria of an RCT, the training sample can be unrepresentative of the testing population we are interested in. Therefore, the corresponding casual evidence may not be broadly applicable or relevant for the real-world practice. In causal inference literature, it is common to regard the training data as a selected sample from the pooled population of training and testing. The selection bias can be adjusted by reweighing or stratifying the training data according to the relationship between training and testing \citep{o2014generalizing,buchanan2018generalizing}. However, it requires strong assumptions on completely measuring the selection confounders and correctly specifying the selection model, and thus can only work well on a prespecified testing population. There are many other practical scenarios where the difference between the training and testing distributions is unknown. One example is that the training data can be confounded by some unidentified effects such as batch effects, which may cause potential covariate shifts \citep{luo2010comparison}. Another possibility is that the testing distribution may evolve over time \citep{hand2006classifier}. There is also a widely studied scenario that multiple datasets are aggregated to perform combined analysis \citep{alyass2015big,shi2018maximin,li2020transfer}. Aggregating data from various sources can benefit from sharing common information, transferring knowledge from different but related samples, and maintaining certain privacy. However, due to the heterogeneity among data sources, standard approaches of finding pooled optimal ITRs may not generalize well on all these sources. One way of handling the heterogeneity is to formulate it as a problem of distributional changes, where we train on the mixture of subpopulations while testing on one of the subpopulations \citep{duchi2019distributionally}. In all these applications, an optimal ITR that is robust to unattended distributional differences is of great interest. 
	
	Despite a vast literature in ITR, much less work has been done on the problem when the training and testing distributions are different. \citet{imai2013estimating} and \citet{johansson2018learning} estimated the CTE function by reweighing the training loss to ensure the estimators generalizable on a prespecified testing distribution. \citet{zhao2019robustifying} aimed to find an ITR that optimizes the worst-case quality assessment among all testing covariate distributions satisfying some moment conditions. However, since their method only requires some moment conditions, the uncertainty set of the testing distributions can be very large. Recent developments in the distributionally robust optimization (DRO) literature provide the opportunities to quantify the difference between the training and testing distributions more precisely \citep{ben2013robust,duchi2018learning,rahimian2019distributionally}. Motivated by the DRO literature, we develop a new robust optimal ITR framework in this paper. 
	
	In this paper, we consider the problem of finding an optimal ITR from a restricted ITR class, where there is some unknown covariate changes between the training and testing distributions. We propose to use the \textit{distributionally robust ITR (DR-ITR)} that maximizes the defined worst-case value function among value functions under a set of underlying distributions. More specifically, value functions are evaluated under all testing covariate distributions that are ``close'' to the training distribution, and the worst-case situation takes a minimal one. Our distributionally robust ITR framework is different from the existing doubly robust ITR framwork that uses an AIPWE. In particular, an AIPWE robustifies the model specification assumptions, while our DR-ITR robustifes the underlying distributions. The DR-ITR aims to guarantee reasonable performance across all testing distributions in an uncertainty set around the training distribution by optimizing the worst-case scenarios. In particular, we parameterize the amount of ``closeness'' by the \textit{distributional robustness-constant (DR-constant)}, where the smallest possible DR-constant corresponds to the \textit{standard ITR} that maximizes the value function under the training distribution. To ensure the performance of the DR-ITR on a specific testing distribution, we fit a class of DR-ITRs for a spectrum of DR-constants at the training stage, and calibrate the DR-constant based on a small amount of the calibrating data from the testing distribution. In this way, the correctly calibrated DR-constant ensures that the DR-ITR performs at least as well as, often much better than, the standard ITR. Using our illustrative example, we show that the standard ITR can have very poor values on many testing distributions, while our calibrated DR-ITRs still maintain relatively good performance. In particular, our proposed calibrating procedures can tune DR-constants based on the small calibrating sample. To solve the worst-case optimization problem, we make use of the difference-of-convex (DC) relaxation of the nonsmooth indicator, and propose two algorithms to solve the related nonconvex optimization problems. We also provide the finite sample regret bound for the proposed DR-ITR.
	
	The rest of this paper is organized as follows. In Section \ref{sec:method}, we discuss an illustrative example that the optimality of an ITR can be sensitive to the underlying distribution, and introduce the DR-ITR that can generalize well across all testing distributions considered in this example. Then we propose the DR-ITR framework and the corresponding learning problem. In Section \ref{sec:theory}, we justify the theoretical guarantees of the finite sample approximations for the learning problem. In Section \ref{sec:simulation}, we evaluate the generalizability of our proposed DR-ITR on two simulation studies: the problem of covariate shifts and the problem of mixture of multiple subgroups. We apply our proposed DR-ITR on the AIDS clinical dataset ACTG 175 and evaluate its generalizability on the subgroup of female patients in Section \ref{sec:aids}. Some related discussions and extensions are given in Section \ref{sec:discuss}. The implementation details, technical proofs and some additional numerical results are all given in the Supplementary Material.

	\section{Methodology} \label{sec:method}
	In this section, we introduce the value maximization framework in the current literature, and discuss its limitation when the training and testing distributions are different. Then we propose the DR-value function that optimizes the worse-case value function across all distributions within an uncertainty set around the training distribution. 
	
	\subsection{Maximizing the Value Function}\label{sec:value_max}
	Consider the training data $ (\bX,A,Y) \sim \bbP $, where $ \bX \in \cX \subseteq \bbR^{p} $ denotes the covariates, $ A \in \cA = \{ +1,-1 \} $ is the binary treatment assignment, and $ Y \in \cY \subseteq \bbR $ is the observed outcome. We assume that the larger outcome is better. Let $ Y(+1),Y(-1) $ be the potential outcomes. Consider a prespecified ITR class $ \cD \subseteq \{ \pm1 \}^{\cX} $. For $ d \in \cD $, denote $ Y(d) := Y(1)\bbone[d(\bX) = 1] + Y(-1)\bbone[d(\bX) = -1] $ as the potential outcome following the treatment assignment prescribed by the ITR $ d $. Then the value function under the training distribution $ \bbP $ is defined as
	\[ \cV(d) := \bbE [Y(d)]. \]
	Denote $ \pi(a|\bx) := \bbP(A = a|\bX = \bx) $ as the training propensity score function for treatment assignment. If we assume 1) the \textit{consistency} of the observed outcome $ Y = Y(A) $; 2) the \textit{strict overlap} $ \pi(\pm 1|\bx) \ge \tau > 0 $ for any $ \bx \in \cX $; and 3) the \textit{strong ignorability} $ (Y(+1),Y(-1)) \indep A|\bX $ \citep{rubin1974estimating}, then we can identify $ \cV(d) $ in terms of the observed data $ (\bX,A,Y) $ by the IPWE of $ \bbE\left( {\bbone[d(\bX)=A] \over \pi(A|\bX)}Y \right) $. 
	
	Instead of targeting the value function directly, we instead consider the CTE function as $ C(\bx) := \bbE[Y(+1) - Y(-1)|\bX=\bx] $ under the training distribution $ \bbP $. Note that for an ITR $ d $ and all $ \bx \in \cX $, the prescribed treatment assignment satisfies $ d(\bx) \in \{ \pm 1 \} $. Then we have $ C(\bx)d(\bx) = \bbE[Y(d) - Y(-d)|\bX = \bx] $. Based on this representation, we define another value function
	\begin{align}
	\cV_{1}(d) := \bbE[C(\bX)d(\bX)] = \bbE[Y(d) - Y(-d)].\label{eq:value_CTE}
	\end{align}
	Since $ Y(d) + Y(-d) \equiv Y(1) + Y(-1) $, it can be observed that $ \cV_{1}(d) = 2\left[ \cV(d) - {\bbE[Y(+1) + Y(-1)] \over 2} \right] = 2[\cV(d) - \cV(d_{\rand})] $, where $ d_{\rand}(\bx) = +1 $ with probability $ 1/2 $ and $ -1 $ with probability $ 1/2 $. Therefore, $ \cV_{1}(d) $ can be interpreted as the value improvement of the ITR $ d $ upon the completely random treatment rule $ d_{\rand} $. In terms of the optimal ITR, the resulting rules by optimizing the value functions $ \cV_{1}(d) $ and $ \cV(d) $ over $ d $ are equivalent. 
	
	By the definition (\ref{eq:value_CTE}), we have $ \cV_{1}(d) \le \bbE [|C(\bX)|] $ with equality if $ d(\bX) = \bsign[C(\bX)] $ almost surely. Such an ITR is the global optimal ITR when $ \cD $ consists of all measurable functions from $ \cX $ to $ \{ \pm 1 \} $. To obtain the global optimal ITR, we can estimate $ C(\bX) $ from data using flexible nonparametric techniques, such as the Bayesian additive regression tree (BART) \citep{hill2011bayesian}, or the casual forest \citep{wager2018estimation}. However, in general, the global optimal ITR $ \bx \mapsto \bsign[C(\bx)] $ can take a very complicated functional form, while decision makers may want to have a simpler ITR \citep{kitagawa2018should}. Then the ITR class $ \cD $ is often considered as a restricted subset of measurable functions from $ \cX $ to $ \{ \pm 1 \} $. The following two-step procedure can be implemented to estimate the restricted optimal ITR on $ \cD $: first we estimate the CTE function $ \bx \mapsto \widehat{C}(\bx) $ using flexible nonparametric techniques; and then we estimate the ITR by solving $ \max_{d \in \cD}\bbE_{n}[\widehat{C}(\bX)d(\bX)] $ on the restricted ITR class $ \cD $ \citep{zhang2012estimating}. Here, $ \bbE_{n} $ is the empirical average based on the training data.
	
	\subsection{Covariate Changes}\label{sec:cov}
	It can be observed that the value functions defined in Section \ref{sec:value_max} depend on the underlying distribution. Suppose we are interested in a testing distribution $ \bbP_{\test} $ that may be different from the training distribution $ \bbP $ to some extent. Then ITRs estimated by most existing methods may not be able to perform well on our target population.	In order to address this problem, we first
	make the following assumption on the potential difference between $ \bbP_{\test} $ and $ \bbP $. 
	\begin{asm}[Covariate Changes]\label{asm:cov} 
		For every training distribution $ \bbP $ and testing distribution $ \bbP_{\test} $ considered in this paper, we assume the followings:
		\begin{itemize}
			\item [(\Romannum{1})] $ \bbP_{\rm test} \ll \bbP $;
			\item [(\Romannum{2})] There exists $ w:\cX \to \bbR_{+} $ such that $ \bbE_{\bbP}w(\bX) = 1 $, and $ \rd\bbP_{\rm test}/\rd \bbP = w(\bX) $.
		\end{itemize}
	\end{asm}
	Assumption \ref{asm:cov} (\Romannum{1}) requires that the support of the testing distribution cannot go beyond the training distribution. Assumption \ref{asm:cov} (\Romannum{2}) is mathematically equivalent to assuming that the differences between $ \bbP $ and $ \bbP_{\rm test} $ only appear in the covariate distributions. The treatment-response relationship conditional on covariates remains unchanged across training and testing distributions. Specifically, let $ p_{\bX}(\bx)p_{Y|\bX}(y(1),y(-1)|\bx) $ and $ q_{\bX}(\bx)q_{Y|\bX}(y(1),y(-1)|\bx) $ be the training and testing densities of the data $ (\bX,Y(1),Y(-1)) $. Then the density ratio $ \rd \bbP_{\test}/\rd \bbP $ becomes
	\[ {\rd \bbP_{\test} \over \rd \bbP} = {q_{\bX}(\bX) \over p_{\bX}(\bX)} \times {q_{Y|\bX}(Y(1),Y(-1)|\bX) \over p_{Y|\bX}(Y(1),Y(-1)|\bX)}. \]
	If $ q_{Y|\bX}(Y(1),Y(-1)|\bX) = p_{Y|\bX}(Y(1),Y(-1)|\bX) $, \textit{i.e.}, the conditional distributions $ (Y(1),Y(-1))|\bX $ are identical under $ \bbP_{\rm test} $ and $ \bbP $, then $ \rd \bbP_{\rm test}/ \rd \bbP = q_{\bX}(\bX)/p_{\bX}(\bX) $, which is the weighting function $ w(\bX) $ in Assumption \ref{asm:cov} (\Romannum{2}). 
	
	The assumption of covariate changes is commonly seen in the setting of randomized trial. Consider the training and testing populations together as a pooled population with finite subjects. For each subject $ i \in \{1,2,\cdots,N\} $, let $ S_{i} \in \{ 0,1 \} $ be a selection random variable such that $ S_{i} = 1 $ if $ i $ is a training sample point, and $ S_{i} = 0 $ if $ i $ is a testing sample point. Let the distributions of $ (\bX_{i},Y_{i}(1),Y_{i}(-1))|(S_{i}=1) $ and $ (\bX_{i},Y_{i}(1),Y_{i}(-1))|(S_{i}=0) $ be the training distribution $ \bbP $ and the testing distribution $ \bbP_{\test} $ respectively. Denote $ \widebar{\bbP} $ as the joint distribution of $ (\bX_{i},Y_{i}(1),Y_{i}(-1),S_{i}) $. Then conditions in Assumption \ref{asm:cov} can correspond to the following \citep{hotz2005predicting,stuart2011use}:
	\begin{itemize}
		\item (Overlapping Support)  $ 0 < \widebar{\bbP}(S_{i}=1|\bX_{i}) < 1 $;
		\item (Selection Unconfoundedness) $ S_{i}\indep (Y_{i}(1),Y_{i}(-1))|\bX_{i} $.
	\end{itemize}
	In particular, under this finite population setting, the overlapping support condition is equivalent to that $ \bbP_{\test} \ll \bbP $ and $ \bbP \ll \bbP_{\test} $, and the selection unconfoundedness condition is equivalent to Assumption \ref{asm:cov} (\Romannum{2}). Such a correspondence can bring more intuitive implications of Assumption \ref{asm:cov} under the randomized trial setting. Specifically, the overlapping support requires the chances of each subject being selected into the training and testing populations to be both positive. The selection unconfoundnedness requires that the selection mechanism is independent of the potential outcomes given the covariates. Both conditions can be satisfied by a successful trial design \citep{pearl2014external}. The phenomenon of covariate changes between $ \bbP $  and $ \bbP_{\test} $ can exist if $ \widebar{\bbP}(S_{i}=1|\bX_{i}) \ne \widebar{\bbP}(S_{i} = 0|\bX_{i}) $ with a positive probability. This can be often the case if the subject needs to satisfy certain requirements before enrolling a trial.

	As a consequence from Assumption \ref{asm:cov}, the CTE function $ C(\bX) = \bbE_{\bbP}[Y(1) - Y(-1)|\bX] = \bbE_{\test}[Y(1) - Y(-1)|\bX] $ remains unchanged under $ \bbP $ and $ \bbP_{\test} $. Then it can be convenient to consider the value functions $ \cV_{1}(d) = \bbE_{\bbP}[C(\bX)d(\bX)] $ and $ \cV_{1,\test}(d) = \bbE_{\test}[C(\bX)d(\bX)] $ defined in (\ref{eq:value_CTE}). When the testing value function $ \cV_{1,\test}(d) $ is of interest, maximizing the training value function $ \cV_{1}(d) $ may not be optimal. Alternatively, we can rewrite the testing value function $ \cV_{1,\test}(d) = \bbE_{\bbP}[w(\bX)C(\bX)d(\bX)] $ where $ w(\bX) = \rd \bbP_{\test}/\rd \bbP $. Then based on the training data from $ \bbP $, we can maximize $ \bbE_{\bbP}[w(\bX)C(\bX)d(\bX)] $ that targets the correct objective. It amounts to determine the weighting function $ w $ that captures the differences between $ \bbP_{\test} $ and $ \bbP $.
	
	\begin{rmk}
		Notice that for any weighting function $ w: \cX \to \bbR_{+} $, we have $ \bbE_{\bbP}[w(\bX)C(\bX)d(\bX)] \le \bbE_{\bbP}[w(\bX)|C(\bX)|] $ with equality if $ d(\bX) = \bsign[C(\bX)] $. That is, if $ \cD $ consists of all measurable functions from $ \cX $ to $ \{ \pm 1 \} $, then the global optimal ITR is \textit{not} sensitive to any covariate changes in the testing distribution. However, the problem of covariate changes induces a challenge if $ \cD $ is a restricted ITR class.
	\end{rmk}
	
	\begin{rmk}
		Our methodology only relies on the fact that $ C(\bX) $ remains unchanged under $ \bbP $ and $ \bbP_{\test} $. Therefore, it can be possible to relax Assumption \ref{asm:cov} to allowing distributional changes in $ (Y(1),Y(-1))|\bX $, while assuming that the CTE function $ C(\cdot) $ remains identical across $ \bbP $ and $ \bbP_{\test} $. Furthermore, our methodology can also be meaningful if the testing CTE function can be different from training, but the optimal treatment assignment remains unchanged. We will discuss this extension in Remark \ref{rmk:cov}. 
	\end{rmk}

	\subsection{An Illustrative Example}\label{sec:eg}

\begin{figure}[!h]
		\centering
		\includegraphics[width=\linewidth]{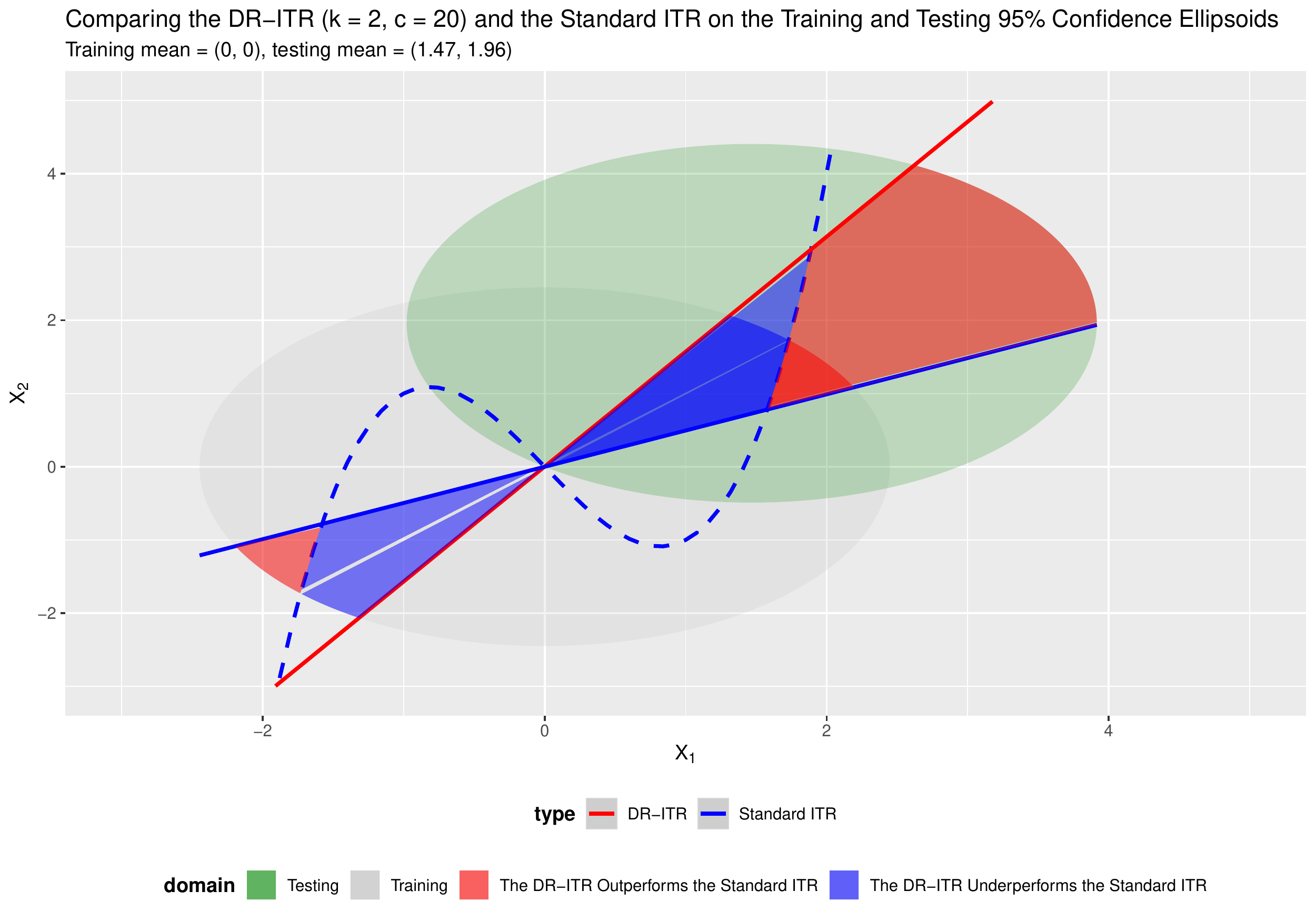}
		\small\caption{ITRs and the 95\% confidence ellipsoids of the training distribution $ (X_{1},X_{2}) \sim \cN_{2}\big((0,0)^{\intercal},\Ib_{2}\big) $ and the testing distribution $ (X_{1},X_{2}) \sim \cN_{2}\big( (1.47,1.96)^{\intercal},\Ib_{2} \big) $. The blue dashed curve is the underlying CTE boundary $ C(X_{1},X_{2}) = X_{2} - (X_{1}^{3} - 2X_{1}) = 0 $.}
		\label{fig:nonlin_par20}
	\end{figure}

	In this section, we begin with an example as in Figure \ref{fig:nonlin_par20} that the optimality of an ITR depends on the underlying distribution. 
	There are two underlying bivariate normal distributions of means $ (0,0)^{\intercal} $ (training) and $ (1.47,1.69)^{\intercal} $ (testing) respectively. We obtain the standard ITR by maximizing the value function $ \cV_{1}(d) $ under the training distribution over the linear ITR class. We also obtain the DR-ITR by maximizing the DR-value function $ \cV^{k}_{c}(d) $ to be introduced in Section \ref{sec:dr_value_max} over the linear ITR class. Then the DR-ITR is compared with the standard ITR through the value functions $ \cV_{1} $ under the training distribution and $ \cV_{1,\test} $ under the testing distribution as in Table \ref{tab:nonlin_par20}. Since the values can be comparable only through the same value function but not across different value functions, we further define the criteria \textit{relative regret} of an ITR as $ [\text{value}(\text{LB-ITR}) - \text{value}(\text{ITR})]/|\text{value}(\text{LB-ITR})| $, where ``value'' can be $ \cV_{1} $ or $ \cV_{1,\test} $, and the LB-ITR maximizes the corresponding value function over the linear ITR class. In this sense, value(LB-ITR) is the best achievable value among the linear ITR class for the corresponding value function, and becomes the benchmark reference for the relative regret criteria.
	
	\begin{table}[!h]
		\centering
		\caption{\label{tab:nonlin_par20}Testing Values (Relative Regrets) Comparisons of ITRs}
		\begin{threeparttable}
			\begin{tabular}{c||c|c||c}
				\toprule
				\diagbox{Value}{ITR}& DR-ITR & Standard ITR & LB-ITR \\
				\midrule
				Training $ \cV_{1} $ & 0.6253 (37.36\%) & 0.9982 (0\%) & 0.9982\\
				\hline
				Testing $ \cV_{1,\test} $ & 4.8230 (9.16\%) & 0.2927 (94.49\%) & 5.3096\\
				\bottomrule
			\end{tabular}
			\begin{tablenotes}
				\fontsize{8}{5}\selectfont
				\item[1] DR-ITR maximizes $ \cV^{k}_{c}(d) $ defined in (\ref{eq:dr_value}) with $ k = 2 $ and $ c = 20 $ over the linear ITR class.
				\item[2] Standard ITR maximizes $ \cV_{1}(d) $ over the linear ITR class.
				\item[3] LB-ITR maximizes $ \cV_{1}(d) $ or $ \cV_{1,\test}(d) $ over the linear ITR class.
				\item[4] Values (larger the better) can be comparable within rows but incomparable between rows.
				\item[5] $ \text{Relative regret}(\text{ITR}) = [\text{value}(\text{LB-ITR}) - \text{value}(\text{ITR})]/|\text{value}(\text{LB-ITR})| $ (smaller the better).
				\item[6] A size-10,000 sample is generated for fitting DR-ITR and LB-ITRs, and an independent size-100,000 sample is generated for evaluation under $ \cV_{1} $ and $ \cV_{1,\test} $.
			\end{tablenotes}
		\end{threeparttable}
	\end{table}
	
	Two facts can be concluded from Table \ref{tab:nonlin_par20}: 1) the optimality of an ITR can be different across different distributions; and 2) maximizing the training value function may have poor testing performance when covariate changes exist. In Table \ref{tab:nonlin_par20}, even though the standard ITR is optimal under the training distribution, it can be far from optimal (94.49\% off in terms of relative regret) under the testing distribution. In contrast, the DR-ITR may not enjoy high training value, but can have much better testing performance (only 9.16\% off in terms of relative regret).
	
	\begin{rmk}\label{rmk:nonlin_par20}
		Figure \ref{fig:nonlin_par20} also illustrates how the covariate changes affect the optimality of ITRs. Specifically, we can divide the covariate domain into two types of subdomains, annotated in blue and red, on which the DR-ITR and standard ITR have different treatment assignments. On the blue subdomain, the standard ITR assignment shares the same sign with the CTE function, while the DR-ITR does not. In this case, the standard ITR outperforms the DR-ITR with the difference of value $ |C(\bX)| $ at the individual level. The case reverses on the red subdomain on which the DR-ITR outperforms the standard ITR. The overall difference of values integrates the individual difference with respect to the training or testing density.
		
		The overall outperformance of the DR-ITR under the testing distribution can be explained from the following three perspectives: 1) the 95\% confidence ellipsoid of the training domain only covers a small area of the red subdomain, while that of the testing domain covers a much larger area; 2) the distance of the red subdomain from the testing centroid is much closer than its distance from the training centroid. Then the red subdomain concentrates higher testing density than training; and 3) the individual value differences $ |C(\bX)| $'s are generally larger on the red subdomain intersected with the testing domain than that intersected with the training domain. Therefore, the DR-ITR performs much better than the standard ITR on the testing distribution.
	\end{rmk}
	
	\subsection{Maximizing the Distributionally Robust Value (DR-Value) Function}\label{sec:dr_value_max}
	
	We begin to introduce our DR-ITR that can show strong generalizability as in Figure \ref{fig:nonlin_par20}. As discussed in Section \ref{sec:intro}, our goal in this paper is not to find an ITR that is generalizable on a specific testing distribution, but rather, to find an ITR that guarantees reasonable performance across an uncertain set of testing distributions. We first define the $ k $-th \textit{power uncertainty set} in two equivalent ways under Assumption \ref{asm:cov}:
	\begin{align}
	\cP^{k}_{c}(\bbP) :&= \left \{ \bbQ \ll \bbP\ \middle|\ \|\rd \bbQ / \rd \bbP\|_{L^{k}(\bbP)} \le c \right \}\label{eq:power}\\
	&= \left \{ \bbQ \ll \bbP\ \middle|\  w: \cX \to \bbR_{+},\ \bbE_{\bbP}w(\bX) = 1,\ \bbE_{\bbP}w(\bX)^{k} \le c^{k},\ {\rd \bbQ \over \rd \bbP} = w(\bX) \right \}. \label{eq:weight}
 	\end{align}
	The set $ \cP^{k}_{c}(\bbP) $ consists of the probability distributions $ \bbQ $ such that the $ L^{k}(\bbP) $-norm of the density ratio $ \rd\bbQ/\rd\bbP $ is bounded above by the DR-constant $ c $. The definition (\ref{eq:weight}) highlights that the density ratio is a weighting function $ w $ of $ \bX $, and the distribution $ \bbQ $ in $ \cP^{k}_{c}(\bbP) $ can be characterized by the weighting function $ w $ satisfying the conditions in (\ref{eq:weight}). Here the DR-constant $ c \ge 1 $ controls the degree of the distributional robustness that measures how ``close'' $ \bbQ $ is from $ \bbP $. In particular, $ c = 1 $ reduces the power uncertainty set $ \cP^{k}_{1}(\bbP) $ to the singleton $ \{ \bbP \} $. The power order $ 1 < k \le +\infty $ parametrizes the measurement of the distance of $ \bbQ $ from $ \bbP $. In particular, the power uncertainty set $ \cP^{k}_{c}(\bbP) $ increases in $ c $ as $ k $ is fixed, and decreases in $ k $ as $ c $ is fixed. The latter one is due to the Lyapunov's inequality: $ \|\rd\bbQ/\rd\bbP\|_{L^{k}(\bbP)} \le \|\rd\bbQ/\rd\bbP\|_{L^{k'}(\bbP)} $ whenever $ 1 < k \le k' \le +\infty $. In the Supplementary Material, we will discuss the explicit form of $ \cP^{k}_{c}(\bbP) $ in the context of specific parametric families of distributions, and how it depends on the DR-constant $ c $ and the power $ k $. One important conclusion from Example S.2 in the Supplementary Material for the mean-shifted $ p $-dimensional normal distribution is that $ \cN_{p}(\bmu,\Ib_{p}) \in \cP^{k}_{c}\big( \cN_{p}(\bzero_{p},\Ib_{p}) \big) $ if and only if $ \|\bmu\|_{2}^{2} \le {2\log c \over k-1} $.
	
	
	With the power uncertainty set $ \cP^{k}_{c}(\bbP) $, we propose to robustly maximize the following worst-case value function among the values under $ \bbQ \in \cP^{k}_{c}(\bbP) $:
	\begin{align}
	\cV^{k}_{c}(d) := \inf_{\bbQ \in \cP^{k}_{c}(\bbP)}\bbE_{\bbQ}[C(\bX)d(\bX)],\label{eq:dr_value}
	\end{align}
	which we term as the \textit{DR-value function}. In particular, $ c = 1 $ reduces the DR-value function $ \cV^{k}_{1}(d) $ to the standard value function $ \cV_{1}(d) = \bbE_{\bbP}[C(\bX)d(\bX)] $ in the definition (\ref{eq:value_CTE}). 
	
	\begin{rmk}[Optimality]
		The ``optimality'' of the DR-ITR is with respect to the DR-value function $ \cV^{k}_{c} $, which highlights its difference from the traditional ``optimal'' ITR with respect to the standard value function $ \cV_{1} $.
	\end{rmk}

\begin{figure}[!htp]
		\centering
		\begin{subfigure}{0.49\linewidth}
			\includegraphics[width=\linewidth]{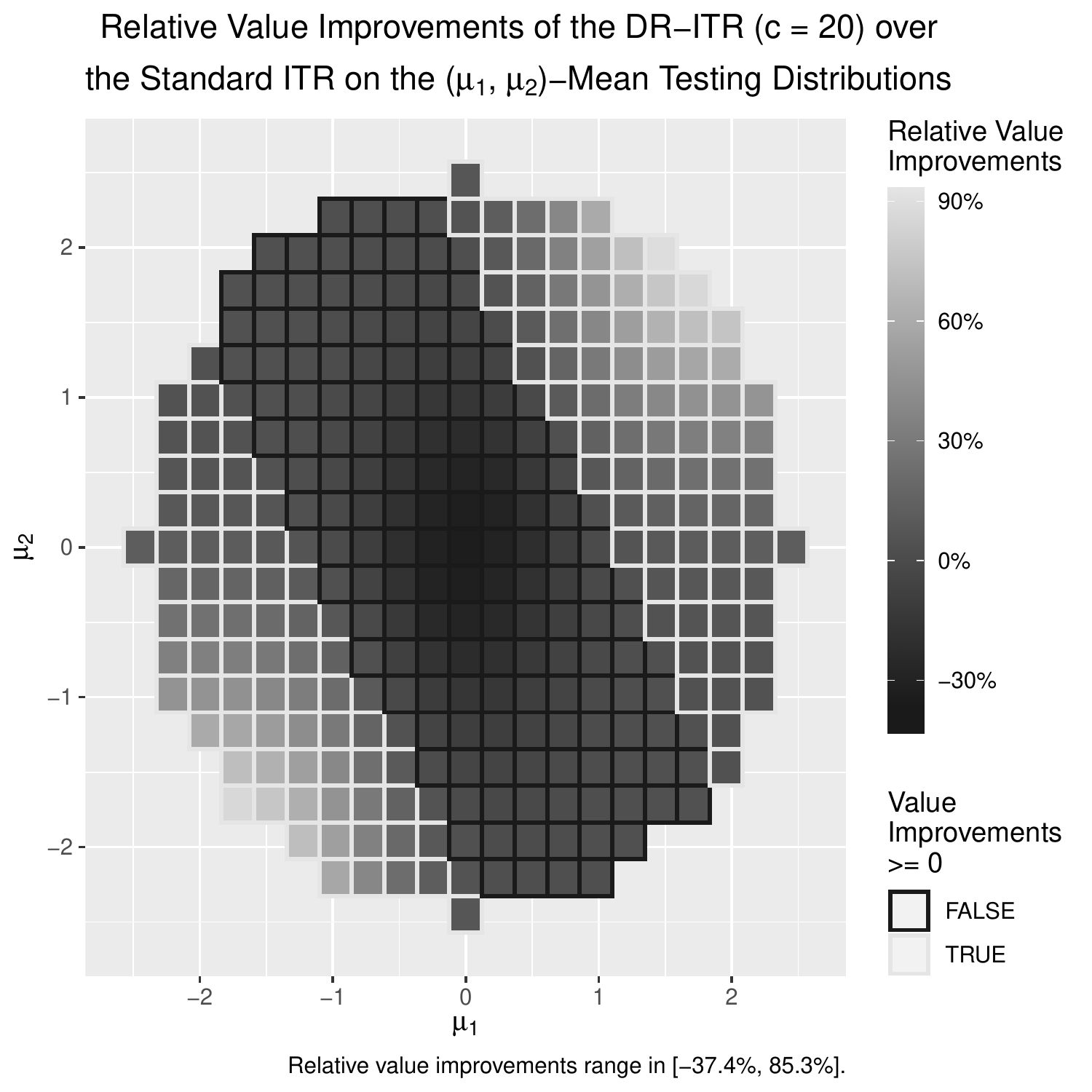}
			\caption{$ c = 20 $}
			\label{fig:shift}
		\end{subfigure}
		\begin{subfigure}{0.49\linewidth}
			\includegraphics[width=\linewidth]{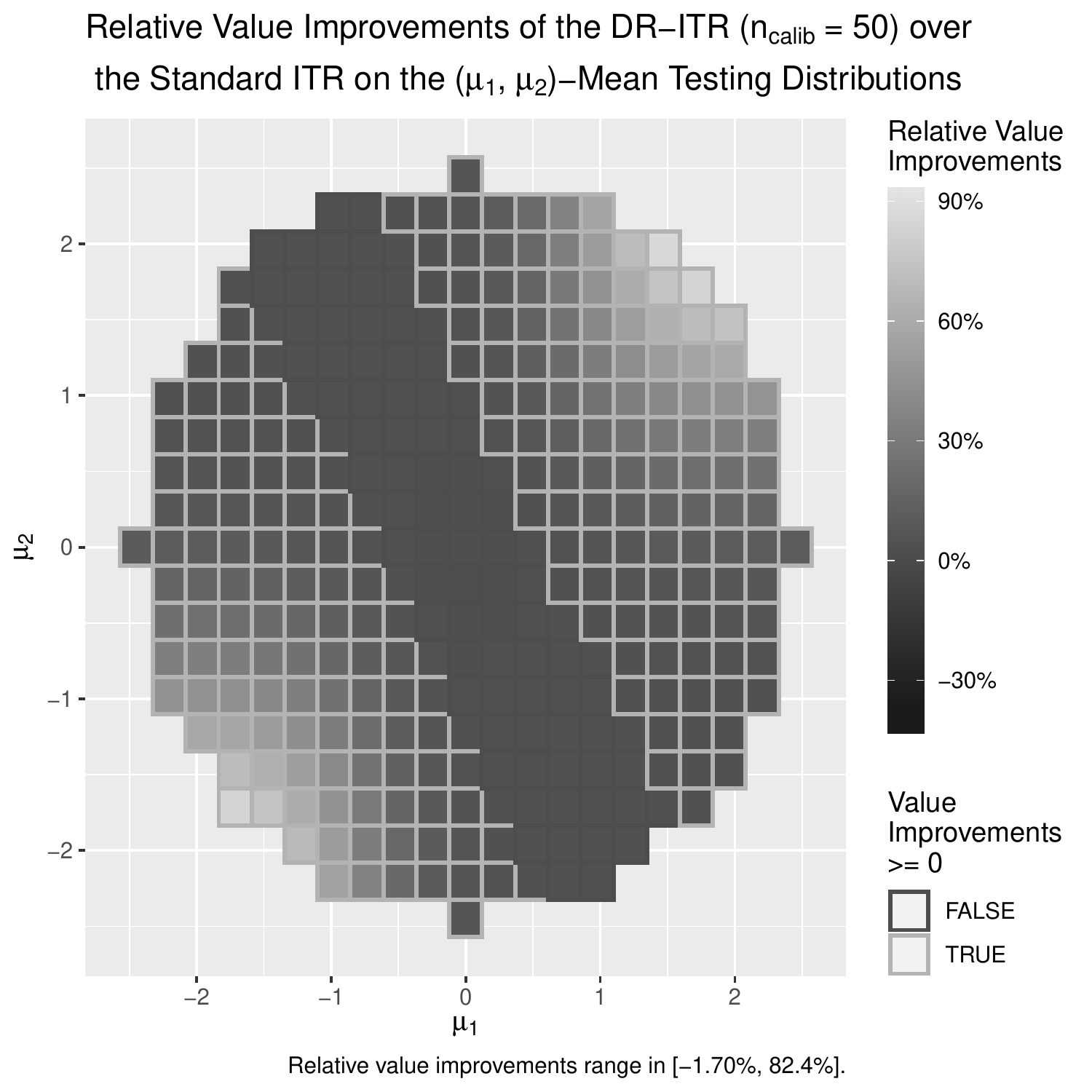}
			\caption{Calibrating $ c $ on a size-50 Sample}
			\label{fig:shift_calib}
		\end{subfigure}
		\caption{Relative improvements of the DR-ITR over the standard ITR as the difference of relative regrets on testing distributions $ \cN_{2}\big(\bmu,\Ib_{2}\big) $ of $ \bmu \in \left \{ (\mu_{1},\mu_{2})^{\intercal} \in \bbR^{2}: \mu_{1}^{2} + \mu_{2}^{2} \le 4\log 5 \right \} $ (lighter the better).}
		\label{fig:nonlin_value_rel_diff}
	\end{figure}

	In the example in Section \ref{sec:eg}, the standard ITR maximizes the value function under the training distribution over the linear ITR class, while the DR-ITR maximizes the DR-value function $ \cV^{k}_{c}(d) $ of $ k = 2 $ and $ c = 20 $ over the linear ITR class. In particular, the randomness of $ \bbP $ comes from the training covariate distribution $ \cN_{2}(\bzero_{2},\Ib_{2}) $. Such a choice of $ \cP^{k}_{c}(\bbP) $ contains the mean-shifted normal distributions $ \cN_{2}(\bmu,\Ib_{2}) $ for all $ \bmu \in \left \{ (\mu_{1},\mu_{2})^{\intercal}: \mu_{1}^{2} + \mu_{2}^{2} \le 4\log 5 \right \} $. In Figure \ref{fig:shift}, we enumerate such mean-shifted normal distributions as the testing distributions, and evaluate the \textit{relative improvement} of the DR-ITR over the standard ITR as the difference of their relative regrets. Among all testing distributions, the relative improvements of the DR-ITR span from $ -37.4\% $ to $ 85.3\% $, suggesting that the potential of improvement can be large. Besides the DR-constant $ c = 20 $, we also consider the case $ c = 2.71,6.51,10.31 $ in the Supplementary Material. As $ c $ increases, the range of relative improvements becomes wider. The increase in the relative improvement upper bound is in general much larger than the decrease in the lower bound.
	
	Based on these observations, the DR-constant $ c $ should be carefully chosen. On one hand, as can be seen from Figure \ref{fig:shift}, the DR-ITR for a fixed DR-constant $ c $ may or may not improve over the standard ITR on a specific testing distribution within $ \cP^{k}_{c}(\bbP) $. When the DR-constant $ c $ can be tuned adaptive to the specific testing distribution, then the DR-ITR can perform at least as well as the standard ITR. On the other hand, we may not even have any prior information on $ c $ to ensure that the power uncertainty set $ \cP^{k}_{c}(\bbP) $ contains the testing distribution of interest. Both cases ask for additional information to calibrate the choice of $ c $ so that the DR-ITR performs well on a specific testing distribution. Suppose we are able to obtain a small size of calibrating sample from the testing distribution. We propose the following training-calibrating procedure to choose $ c $: 1) at the training stage, we estimate DR-ITRs $ \{ \widehat{d}_{c} \}_{c \in \cC} $ where $ c $ is the DR-constant to compute $ \widehat{d}_{c} $, and $ \cC $ is a set of candidate DR-constants; 2) we obtain a calibrating sample from the testing distribution, on which we estimate the testing values of $ \{ \widehat{d}_{c} \}_{c\in \cC} $; 3) we select the $ \widehat{c} $ that maximizes the value of $\widehat{d}_{c} $ among $ c \in \cC $. 
	
	In order to estimate the value function under the testing distribution, we consider the following two possible calibration scenarios: 1) the calibrating sample is a randomized controlled trial (RCT) dataset $ (\bX,A,Y) $ from the testing distribution; and 2) the calibrating sample only consists of the covariates $ \bX $ from the testing distribution. Scenario 1 will be more ideal than Scenario 2 since we have the testing information of both the treatment and the outcome. We can evaluate an ITR $ d $ using the IPWE $ \widehat{\cV}_{\rm calib}^{\IPWE}(d) = \bbE_{n_{\rm calib}}\{ \bbone[d(\bX) = A]Y/\pi_{\rm calib}(A|\bX) \} $, where $ \bbE_{n_{\rm calib}} $ is the empirical average over the calibrating sample, $ \pi_{\rm calib} $ is the corresponding propensity score function, and $ \pi_{\rm calib} $ is known or estimable from the calibrating data. We call the corresponding calibrate DR-ITR as \textit{RCT-DR-ITR}. In Scenario 2, we do not have the treatment-response information from the testing distribution. We can instead use the value function estimate $ \widehat{\cV}_{\rm calib}^{\CTE}(d) = \bbE_{n_{\rm calib}}[\widehat{C}_{n}(\bX)d(\bX)] $ to evaluate $ d $, where $ \widehat{C}_{n}(\bX) $ is estimated at the training stage. However, the CTE estimate $ \widehat{C}_{n}(\cdot) $ may also suffer from a potential generalizability problem on the testing distribution.  Practitioners need to be careful of the generalizability of the CTE estimate when performing the calibration. We call the corresponding DR-ITR as \textit{CTE-DR-ITR}. 
	
	RCT-DR-ITR and CTE-DR-ITR are different in their use of information for calibration. Specifically, the RCT-DR-ITR makes use of $ (\bX,A,Y) $ from the testing distribution, while the CTE-DR-ITR only makes use of $ \bX $ from the testing distribution, and the underlying CTE function $ C(\bX) $. In practice, $ C(\bX) $ is estimated from training data. It requires Assumption \ref{asm:cov} to generalize the CTE estimate $ \widehat{C}_{n}(\bX) $ from training to testing. If Assumption \ref{asm:cov} holds, then CTE-DR-ITR can have better performance than RCT-DR-ITR, since CTE-DR-ITR captures less variance from calibrated data. If Assumption \ref{asm:cov} is violated, which will be illustrated in Section \ref{sec:mix}, then CTE-DR-ITR can have poorer performance than RCT-DR-ITR, since the testing value function estimate of CTE-DR-ITR can be biased.
	
	In Figure \ref{fig:shift_calib}, we generate a calibrating RCT sample from $ \bbP_{\test} $ of size 50. It shows that across the mean-shifted testing distributions, the relative improvements of the calibrated DR-ITRs range from $-1.70\%$ to $82.4\% $. It suggests that the small sample size 50 is sufficient for a reasonably good calibration, with the positive relative improvements being maintained.

	\begin{rmk}[Extending Covariate Changes]\label{rmk:cov}
		Consider the case that Assumption \ref{asm:cov} is violated. Let $ C_{\test} $ be the testing CTE function that can be different from the training CTE function $ C $. We use the notations $ \bbP $ and $ \bbP_{\test} $ to refer to the training and testing covariate distributions. Assume that $ \bsign[C_{\test}(\bX)] = \bsign[C(\bX)] $ almost surely. Then we can still represent the value function under the testing distribution as follows:
		\[ \bbE_{\test}[C_{\test}(\bX)d(\bX)] = \bbE_{\bbP}\left\{ {\rd \bbP_{\test} \over \rd \bbP} {C_{\test}(\bX) \over C(\bX)}\bbone[C(\bX) \ne 0] \times C(\bX)d(\bX) \right\}. \]
		The definition of the DR-value function (\ref{eq:dr_value}) can be robust with respect to the change of $ (\bbP_{\test},C_{\test}) $ from $ (\bbP,C) $, such that $ w(\bX) := (\rd \bbP_{\test}/\rd \bbP) \times [C_{\test}(\bX)/C(\bX)]\bbone[C(\bX) \ne 0] $ satisfies $ \bbE_{\bbP}w(\bX) = 1 $ and $ \bbE_{\bbP}w(\bX)^{k} \le c^{k} $.
	\end{rmk}

	\begin{rmk}
		The calibration procedure ensures that among the DR-ITRs of various DR-constants, the best one is chosen to maximize the testing value function. In this sense, the calibrated DR-ITR can have potential of improving the generalizability from training to testing. However, if the testing distribution is very far from the training distribution, one cannot expect that an ITR estimated by any method from the training data can perform well on the test data, even though our proposed method may be able to protect against such a distributional change to some extent. Therefore, in practice, we suggest to use our method when training and testing distributions are relatively close.
	\end{rmk}
	
	\subsection{Distributionally Robust Expectation} \label{sec:dre} 
	In this section, we first discuss the rationale of considering the $ L^{k} $-norm of the density ratio as the measurement of distributional distance. We show that the $ k $-th power uncertainty set $ \cP^{k}_{c}(\bbP) $ is equivalent to the distributional ball induced by the $ \phi $-divergence \citep{pardo2005statistical} for some specific divergence $ \phi $. Then we derive the dual form of the worst-case expectation over $ \cP^{k}_{c}(\bbP) $, which provides a more tractable optimization problem.
	
	\subsubsection{Equivalence to the Divergence-Based Distributional Ball}\label{sec:divg}
	As a generalization of the conventional likelihood-based framework which corresponds to the Kullback-Leibler (KL) divergence, the framework of general $ \phi $-divergence between distributions has been well studied in the context of parameter estimation and hypothesis testing  \citep{pardo2005statistical}. The $ \phi $-divergence between two probability distributions $ \bbP $ and $ \bbQ $ such that $ \bbQ \ll \bbP $ is defined as follows:
	\[ D_{\phi}(\bbQ\|\bbP) := \int \phi\left( {\rd \bbQ \over \rd \bbP} \right) \rd \bbP = \bbE_{\bbP}\phi\left( {\rd \bbQ \over \rd \bbP} \right); \quad\phi \in \Phi, \]
	where $ \Phi $ is a class of convex functions on $ \bbR $ that satisfies the regularity conditions: $ \phi(w) = +\infty $ for $ w < 0 $, $ \phi(1) = \phi'(1) = 0 $, and $ \lim\limits_{w \to 0_{+}}w\phi(p/w) = \lim\limits_{w \to +\infty}\phi(w)/w $ for $ p > 0 $. The definition with various choices of $ \phi $'s includes the empirical likelihood $ \phi_{\EL}(w) = - \log w + w - 1 $, the KL divergence $ \phi_{\KL}(w) = w\log w - w + 1 $, and the $ \chi^{2} $-divergence $ \phi_{\chi^{2}}(w) = {1 \over 2}(w - 1)^{2} $. There is another important special case that relates to the power uncertainty set of $ k = +\infty $. Consider the optimization indicator for $ c \ge 1 $: $ \phi_{\infty,c} = 0 $ if $ u \in [0,c] $ and $ +\infty $ otherwise, for which $ D_{\phi_{\infty,c}}(\bbQ \|\bbP) = 0 \text{ if } \left\| \rd\bbQ/\rd\bbP \right\|_{L^{\infty}(\bbP)} \le c $, and $ +\infty $ otherwise. Then $ D_{\phi_{\infty,c}}(\bbQ\|\bbP) = 0 $ if and only if $ \bbQ \in \cP^{\infty}_{c}(\bbP) $.
	
	Although $ D_{\phi} $ is not a proper metric between probability distributions since it is asymmetric, we can still define a $ D_{\phi} $-distributional ball as $ \sP^{\phi}_{\rho}(\bbP) := \{ \bbQ\ll\bbP: D_{\phi}(\bbQ\|\bbP) \le \rho \} $, where $ \bbP $ is the center and $ \rho \ge 0 $ is the radius. Then for any $ \rho \ge 0 $, the $ D_{\phi_{\infty,c}} $-distributional ball $ \sP^{\phi_{\infty,c}}_{\rho}(\bbP) \equiv \{ \bbQ \ll \bbP: D_{\phi_{\infty,c}}(\bbQ\|\bbP) = 0 \} $, which coincides with the power uncertainty set $ \cP^{\infty}_{c}(\bbP) $ defined in (\ref{eq:power}) for $ k = \infty $. Such an equivalence can be extended to all finite $ k \in (1,+\infty) $ when a Cressie-Read (CR) family \citep{cressie1984multinomial} of divergence functions $ \Phi_{\CR} \subseteq \Phi $ is taken into consideration. For $ k > 1 $, the corresponding $ \phi_{k} \in \Phi_{\CR} $ is defined as
	\[ \phi_{k}(w) := {w^{k} - kw + k - 1 \over k(k-1)}; \quad w \ge 0. \]
	Here, $ \phi_{k} $ effectively measures the probability-distributional distance by the $ k $-th moment of the density ratio, since $ D_{\phi_{k}}(\bbQ\|\bbP) = {1 \over k(k-1)}[\bbE_{\bbP}(\rd \bbQ/ \rd \bbP)^{k} - 1] $ as long as $ \bbQ $ is a probability distribution. Then it can be inferred that the $ D_{\phi_{k}} $-distributional ball $ \sP^{\phi_{k}}_{\rho}(\bbP) $ is actually equivalent to the power uncertainty set $ \cP^{k}_{c_{k}(\rho)}(\bbP) $ in (\ref{eq:power}). Here, there is a one-to-one correspondence between the DR-constant $ c $ and the radius $ \rho $ of the $ D_{\phi_{k}} $-distributional ball with $ c_{k}(\rho) := [k(k-1)\rho + 1]^{1/k} $. We conclude the case $ k = +\infty $ and $ 1 < k < +\infty $ with the following:
	\begin{align}
	\sP^{\phi_{\infty,c}}_{\rho}(\bbP) = \cP^{\infty}_{c}(\bbP); \quad \sP^{\phi_{k}}_{\rho}(\bbP) = \cP^{k}_{c_{k}(\rho)}(\bbP); \quad \rho \ge 0.\label{eq:divg_set}
	\end{align}
	
	\subsubsection{Dual Representation}\label{sec:dual}
	We begin with a general result on the dual representation of the $ \phi $-divergence-based distributionally robust expectation. We state the following lemma and refer readers to \citet[Proposition 1]{duchi2018learning}.
	
	\begin{lem}\label{lem:dual}
		Fix a random variable $ Z $ on $ \bbR $ with distribution $ \bbP $. Let $ \phi \in \Phi $ be a legitimate divergence function. Define the convex conjugate of $ \phi $ as
		\[ \phi^{\star}(x^{\star}) := \sup_{x \in \bbR}\{\langle x^{\star},x \rangle - \phi(x)\}; \quad x^{\star} \in \bbR. \]
		Then for $ \rho > 0 $,
		\begin{align}
		\sup_{\bbQ \in \sP^{\phi}_{\rho}(\bbP)}\bbE_{\bbQ}Z = \inf_{\substack{\lambda \ge 0 \\ \eta \in \bbR}}\left\{\bbE_{\bbP}\left[ \lambda \phi^{\star}\left( {Z - \eta \over \lambda} \right) \right] + \lambda \rho + \eta\right\}.\label{eq:dual}
		\end{align}
	\end{lem}
	
	Let $ c \ge 1 $. Lemma \ref{lem:dual} can be directly applied to the optimization indicator: $ \phi_{\infty,c}(u) := 0 $ if $ u \in [0,c] $ and $ +\infty $ otherwise, whose convex conjugate is given by $ \phi^{\star}_{\infty,c}(u) = c\max\{u,0\} $. Then $ \lambda $ in (\ref{eq:dual}) attains the infimum at $ \lambda = 0 $, so that
	\begin{align}
	\sup_{\bbQ \in \sP^{\phi_{\infty,c}}_{\rho}(\bbP)}\bbE_{\bbQ}Z = \inf_{\eta \in \bbR}\left\{ c\bbE_{\bbP}(Z - \eta)_{+} + \eta \right\}.\label{eq:cvar}
	\end{align}
	In particular, the right hand side of (\ref{eq:cvar}) is solved by the $ (1-1/c) $-\textit{value-at-risk $ \VaR_{1-1/c} $} in finance, or equivalently, the $ (1-1/c) $-quantile of $ Z $ under the center distribution $ \bbP $. The right hand side of (\ref{eq:cvar}) itself is defined as the $ (1-1/c) $-\textit{conditional value-at-risk $ \CVaR_{1-1/c} $} \citep{rockafellar2000optimization}. Next, we apply Lemma \ref{lem:dual} to the $ k $-th power divergence $ \phi_{k} $ to derive the dual problem of the worst-case expectation over $ \cP^{k}_{c}(\bbP) $.
	
	\begin{lem}\label{lem:dual_cr}
		Let $ \Phi_{\CR} $ be the Cressie-Read family of divergence functions, $ k,k^{\star} \in (1,+\infty) $ be conjugate numbers, \textit{i.e.}, $ {1 \over k} + {1 \over k^{\star}} = 1 $, and $ \phi_{k} \in \Phi_{\CR} $. Then we have following conclusions:
		\begin{itemize}
			\item [(\Romannum{1})] The convex conjugate of $ \phi_{k} $ is given by
			\[ \phi^{\star}_{k}(z) = {1 \over k}\left\{ [(k-1)z + 1]^{k^{\star}}_{+} - 1 \right\}. \]
			
			\item [(\Romannum{2})] Fix a probability measure $ \bbP $ and a random variable $ Z $ on $ \bbR $. Then for $ \rho \ge 0 $,
			\begin{align}
			\sup_{\bbQ \in \sP^{\phi_{k}}_{\rho}(\bbP)}\bbE_{\bbQ}Z = \inf_{\eta\in\bbR}\left\{ c_{k}(\rho)[\bbE_{\bbP}(Z - \eta)_{+}^{k^{\star}}]^{1/k^{\star}} + \eta\right\},\label{eq:dual_cr}
			\end{align}
			where $ c_{k}(\rho) = [k(k-1)\rho + 1]^{1/k} $.
		\end{itemize}
	\end{lem}

	Note that the right hand side of (\ref{eq:dual_cr}) and its optimizer $ \eta $ are both coherent risk measures as the higher-order generalizations of the CVaR and VaR \citep{krokhmal2007higher}. 
	
	Using the equivalence in (\ref{eq:divg_set}), the worst-case expectation over the power uncertainty set $ \cP^{k}_{c}(\bbP) $ for $ k \in (1,\infty] $ and $ k^{\star} = {k \over k - 1} $ (in particular, $ k = \infty $ $ \Leftrightarrow $ $ k^{\star} = 1 $) unifies (\ref{eq:cvar}) and (\ref{eq:dual_cr}) as follows:
	\begin{align}
	\sup_{\bbQ \in \cP_{c}^{k}(\bbP)}\bbE_{\bbQ}Z = \inf_{\eta \in \bbR}\left\{ c[\bbE_{\bbP}(Z-\eta)^{k^{\star}}_{+}]^{1/k^{\star}} + \eta \right\}; \quad c \ge 1. \label{eq:dr_dual}
	\end{align}
	By inspecting the dual problem (\ref{eq:dr_dual}), the right hand side is computationally more tractable than the left hand side, since instead of optimizing over an infinite-dimensional probability measure $ \bbQ $, we only need to optimize over a univariate variable $ \eta $. 

	In order to apply the duality result to the DR-ITR problem, we negate the DR-value maximization to a risk minimization problem. Denote the \textit{risk function} under the training distribution $ \bbP $ as $ \cR_{1}(d) := -\cV_{1}(d) = \bbE_{\bbP}\{C(\bX)[-d(\bX)]\} $. Then for $ k \in (1,+\infty]$ and $ c\ge 1 $, the \textit{DR-risk function} is defined as
	\[ \cR^{k}_{c}(d) := \sup_{\bbQ \in \cP^{k}_{c}(\bbP)}\bbE_{\bbQ}\{C(\bX)[-d(\bX)]\}. \]
	Using the fact $ Z = -C(\bX)d(\bX) = C(\bX) \bbone[d(\bX) = -1] + [-C(\bX)]\bbone[d(\bX) = 1] $, the dual representation (\ref{eq:dr_dual}) can be expressed in the following particular form (\ref{eq:dritr_dual}).
	
	\begin{coro}[Dual Representation of the DR-Risk Function]\label{coro:risk_dual}
		Let $ k \in (1,+\infty] $, $ k^{\star} = {k \over k-1} $ if $ k < +\infty $ and $ k^{\star} = 1 $ if $ k = +\infty $, $ c \ge 1 $. Then the DR-risk function $ \cR_{c}^{k} $ has the following dual representation:
		\begin{align}
		\cR^{k}_{c}(d) = \inf\limits_{\eta \in \bbR} \left\{ c\left[ \bbE\left( [C(\bX) - \eta]_{+}^{k^{\star}}\bbone[d(\bX) = -1] + [-C(\bX) - \eta]_{+}^{k^{\star}}\bbone[d(\bX) = 1]\right) \right]^{1/k^{\star}} + \eta\right\}. \label{eq:dritr_dual}
		\end{align}	
	\end{coro}

	\subsection{Implementation} \label{sec:implement}
	In this section, we introduce the implementation of DR-risk minimization based on the empirical data. We cast the learning problem as finding a decision function $ f: \cX \to \bbR $ that induces an ITR based on its sign: $ d(\bx) = \bsign[f(\bx)] $. The ITR class $ \cD $ can correspond to a prespecified decision function class $ \cF $. The DR-risk function as a functional of the decision function becomes $ \cR^{k}_{c}(f) = \sup_{\bbQ \in \cP^{k}_{c}(\bbP)}\bbE_{\bbQ}\big\{C(\bX)\bsign[-f(\bX)]\big\} $. However, directly optimizing the risk $ \cR^{k}_{c}(f) $ is challenging, since the $ \bsign(\cdot) $ operation is nonconvex and nonsmooth. We consider a specific difference-of-convex (DC) relaxation of the sign operator.
	
	We propose to relax the indicators in the dual form (\ref{eq:dritr_dual}) by the following robust smoothed ramp loss \citep{zhou2017residual}: $ \psi(u) := (1-u)^{2}\bbone(0 \le u \le 1) + [2 - (1+u)^{2}]\bbone(-1 \le u \le 0) + 2\bbone(u \le -1) $. The DC representation is given by $ \psi(u) = \psi_{+}(u) - \psi_{-}(u) $, where $ \psi_{+}(u) = (1-u)^{2}\bbone(0 \le u \le 1) + (1-2u)\bbone(u \le 0) $, $ \psi_{-}(u) = u^{2} \bbone(-1 \le u \le 0) + (-1-2u)\bbone(u \le -1) $. The advantages of using the symmetric nonconvex loss can be: 1) to protect from outliers in $ \bX $ and improve generalizability \citep{shen2003psi,wu2007robust}, and 2) to equally indicate $ f(\bX) < 0 $ and $ f(\bX) > 0 $. We would like to point out that $ \bbone[f(\bX) < 0] + \bbone[f(\bX) > 0] \equiv 1 $ will be preserved to $ {\psi[f(\bX)] \over 2} + {\psi[-f(\bX)] \over 2} \equiv 1 $ in this surrogate loss. Then we define the DR-$ \psi $-risk function as
	\begin{align}
	\cR^{k}_{c,\psi}(f) &:= \inf_{\eta \in \bbR}\left\{ c\left[\bbE\left(  [C(\bX) - \eta]_{+}^{k^{\star}}{\psi[f(\bX)]\over 2} + [-C(\bX) - \eta]_{+}^{k^{\star}}{\psi[-f(\bX)] \over 2}\right) \right]^{1/k^{\star}} + \eta\right\}. \label{eq:dritr_risk}
	\end{align}
	
	Algebraically, we can invert (\ref{eq:dritr_risk}) to its primal representation $ \cR^{k}_{c,\psi}(f) = \sup_{\bbQ \in \cP^{k}_{c}(\bbP)}\bbE_{\bbQ}[C(\bX)\zeta_{\psi}(f)] $ by introducing a sign random variable $ \zeta_{\psi}(f) \in \{ \pm 1 \} $ with $ \bbP(\zeta_{\psi}(f) = \pm 1|\bX) := {\psi[\pm f(\bX)] \over 2} $. That is, given the covariate $ \bX $, the original deterministic sign $ \bsign[-f(\bX)] $ is relaxed to the random sign $ \zeta_{\psi}(f) $ with $ \pm 1 $ probability $ {\psi[\pm f(\bX)]\over 2} $. In particular, if $ f(\bX) > 0 $, then $ \bsign[-f(\bX)] = -1 $ is a hard sign while $ \zeta_{\psi}(f) $ is a soft sign with $ \bbP(\zeta_{\psi}(f) = -1|\bX) = {\psi[-f(\bX)] \over 2} > {\psi[f(\bX)] \over 2} = \bbP(\zeta_{\psi}(f) = 1|\bX) $. When $ c = 1 $, the DR-risk function reduces to the risk function under the training distribution, and the DC relaxation here is equivalent to the relaxation in \citet{zhou2017residual}. 
	
	The DR-$ \psi $-risk function provides the learning objective based on the empirical data. In particular, the population expectation $ \bbE $ is replaced by the empirical average $ \bbE_{n} $, and the CTE function $ C(\cdot) $ is replaced by a plug-in estimate $ \widehat{C}_{n}(\cdot) $. The corresponding empirical objective is minimized over the decision function $ f $ and the auxiliary variables $ (\eta,\lambda) $ jointly:
	\begin{align*}
	\min_{f\in\cF,\eta\in\bbR}&\left\{ c\left[ \bbE_{n}\left( [\widehat{C}_{n}(\bX) - \eta]^{k^{\star}}_{+}{\psi[f(\bX)] \over 2} + [-\widehat{C}_{n}(\bX) - \eta]^{k^{\star}}_{+}{\psi[-f(\bX)] \over 2} \right) \right]^{1/k^{\star}} + \eta\right\}\\
	=\min_{f\in\cF,\eta\in\bbR,\lambda \ge 0}&\left\{ {c \over k^{\star}\lambda^{k^{\star}-1}}\bbE_{n}\left( [\widehat{C}_{n}(\bX) - \eta]^{k^{\star}}_{+}{\psi[f(\bX)] \over 2} + [-\widehat{C}_{n}(\bX) - \eta]^{k^{\star}}_{+}{\psi[-f(\bX)] \over 2} \right) + {c\lambda \over k} + \eta\right\}.
	\end{align*}
	The objective function is a summation of multiple products of DC functions. For $ k < +\infty $, we consider a block successive upper-bound minimization algorithm \citep{razaviyayn2013unified} to alternatively minimize the convex upper bounds over the decision function $ f $ and the auxiliary variables $ (\eta,\lambda) $ respectively. For $ k = +\infty $, it requires a further probabilistic enhancement to break ties at argmin and ensure the convergence to stationarity \citep{qi2019estimation,qi2019estimating}. The implementation details are given in the Supplementary Material.

	\section{Theoretical Properties} \label{sec:theory}
	In this section, we justify the validity of the DC relaxation and the empirical substitution. First of all, we introduce the following joint stochastic objectives:
	\begin{align*}
	\ell^{k}_{c}(f,\eta,\lambda;\widetilde{C}) &:= {c \over k^{\star}\lambda^{k^{\star}-1}}\left( [\widetilde{C}(\bX) - \eta]_{+}^{k^{\star}}\bbone[f(\bX) < 0] + [-\widetilde{C}(\bX) - \eta]_{+}^{k^{\star}}\bbone[f(\bX) > 0]\right) + {c\lambda \over k} + \eta;\\
	\ell^{k}_{c,\psi}(f,\eta,\lambda;\widetilde{C}) &:= {c \over k^{\star}\lambda^{k^{\star}-1}}\left( [\widetilde{C}(\bX) - \eta]_{+}^{k^{\star}}{\psi[f(\bX)] \over 2} + [-\widetilde{C}(\bX) - \eta]_{+}^{k^{\star}}{\psi[-f(\bX)] \over 2}\right) + {c\lambda \over k} + \eta.	
	\end{align*}
	Here, $ \widetilde{C} $ can be the plug-in estimate $ \widehat{C}_{n} $ or the underlying true CTE $ C $. Denote $ \cL^{k}_{c}(f,\eta,\lambda) := \bbE\ell^{k}_{c}(f,\eta,\lambda;C) $, $ \cL^{k}_{c,\psi}(f,\eta,\lambda) := \bbE\ell^{k}_{c,\psi}(f,\eta,\lambda;C) $. Then by Corollary \ref{coro:risk_dual}, we have $ \cR^{k}_{c}(f) = \inf_{\eta \in \bbR, \lambda \ge 0}\cL^{k}_{c}(f,\eta,\lambda) $, $ \cR^{k}_{c,\psi}(f) = \inf_{\eta \in \bbR, \lambda \ge 0}\cL^{k}_{c,\psi}(f,\eta,\lambda) $. In the following proposition, we show the validity of the DC relaxation.
		
	\begin{prop}[Fisher Consistency and Excess Risk]\label{prop:fisher} Suppose $ \cR^{k}_{c} $, $ \cR^{k}_{c,\psi} $, $ \cL^{k}_{c} $ and $ \cL^{k}_{c,\psi} $ are defined as above. Fix $ k \in (1,+\infty] $, $ k^{\star} = {k \over k-1} $, $ c \ge 1 $, $ \eta \in \bbR $, $ \lambda > 0 $. Then the following results hold:
	\begin{enumerate}[label=(\Roman*)]
		\item (Fisher Consistency) 
		\[ \argmin\limits_{f:\cX \to [-1,1]}\cL^{k}_{c,\psi}(f,\eta,\lambda) = \argmin\limits_{f:\cX \to \{ \pm 1 \}}\cL^{k}_{c}(f,\eta,\lambda), \quad \min\limits_{f:\cX \to [-1,1]}\cL_{c,\psi}^{k}(f,\eta,\lambda) = \min\limits_{f:\cX \to \{ \pm 1 \}}\cL^{k}_{c}(f,\eta,\lambda); \]
		\item (Excess Risk) Denote $ \cL^{k,*}_{c}(\eta,\lambda) := \min_{f \in \cX \to \{ \pm 1 \}}\cL^{k}_{c}(f,\eta,\lambda) $. Then for $ f:\cX \to \bbR $, we have 
		\[ \cL^{k}_{c}(f,\eta,\lambda) - \cL^{k,*}_{c}(\eta,\lambda) \le 2[\cL^{k}_{c,\psi}(f,\eta,\lambda) - \cL^{k,*}_{c}(\eta,\lambda)]. \]
		Denote $ \cR^{k,*}_{c} := \inf_{\eta \in \bbR, \lambda \ge 0}\cL^{k,*}_{c}(\eta,\lambda) $. Then for $ f: \cX \to \bbR $, we have 
		\[ \cL^{k}_{c}(f,\eta,\lambda) - \cR^{k,*}_{c} \le 2[\cL_{c,\psi}^{k}(f,\eta,\lambda) - \cR^{k,*}_{c}], \quad \cR^{k}_{c}(f) - \cR^{k,*}_{c} \le 2[\cR^{k}_{c,\psi}(f) - \cR^{k,*}_{c}]. \]
	\end{enumerate}
	\end{prop}
	
	Suppose $ \cF $ is a functional class on $ \cX $ with norm $ \|\cdot\|_{\cF} $ that characterizes the complexity of function. Motivated by  \citet[(6)]{steinwart2007fast}, we define for $ \gamma \ge 0 $ the constrained version of the approximation error
	\[ \cA_{c}^{k}(\gamma) := \inf_{f \in \cF}\left\{ \cR_{c,\psi}^{k}(f): \|f\|_{\cF} \le \gamma \right\} - \cR^{k,*}_{c}. \]
	Similarly to that in \citet{steinwart2007fast}, $ \cA^{k}_{c}(\gamma) $ with the appropriately chosen tuning parameter $ \gamma $ can trade off the learnability and the approximatability of $ \cF $ towards the population Bayes rule $ \argmin_{f: \cX \to \{ \pm 1 \}} \cR_{c}^{k}(f) $. Specifically, as $ \gamma $ increases, the population approximation error (``bias'') $ \cA^{k}_{c}(\gamma) $ decreases with $ \gamma $, while the empirical complexity (``variance'') increases with $ \gamma $. The trade-off will be stated more explicitly in the following Assumption \ref{asm:prox}.
	
	Next, we make the following assumptions to show the regret bound for the empirical minimization of the $ \psi $-risk $ \bbE_{n}\ell^{k}_{c,\psi}(f,\eta,\lambda;\widehat{C}_{n}) $. Without loss of generality, we restrict to consider the functional class $ \cF $ as the Reproducing Kernel Hilbert Space (RKHS) with the Gaussian radial basis function kernels, where $ \|\cdot\|_{\cF} $ is the RKHS-norm. General results can be established by adopting the covering number argument as in \citet[Theorem 3.1]{zhao2019efficient}. 

	\begin{asm}[Boundedness]\label{asm:bdd} There exists $ M < +\infty $ such that $ |C(\bX)| \le M $ almost surely.
	\end{asm}

	\begin{asm}[Diffuse Property]\label{asm:diff}
		The distribution of $ C(\bX) $ has a uniformly bounded density with respect to the Lebesgue measure.
	\end{asm}

	\begin{asm}[Convergence of the Plug-in CTE]\label{asm:cte}
		For the CTE estimate $ \widehat{C}_{n}(\bX) $, we assume that
		$ \|\widehat{C}_{n} - C\|_{\infty} := \sup\limits_{\bx \in \cX}\left| \widehat{C}_{n}(\bx) - C(\bx) \right| \overset{\bbP}{\to} 0 $.
	\end{asm}

	\begin{asm}[Approximation Error Rate]\label{asm:prox}
		There exists $ \beta \in (0,1] $ and $ K_{\cA} < + \infty $ such that for all small enough $ \gamma > 0 $, we have $ \cA^{k}_{c}(\gamma) \le K_{\cA}\gamma^{-\beta} $.
	\end{asm}

	As a remark, we note that Assumption \ref{asm:bdd} can hold if the difference of potential outcomes $ Y(1) - Y(-1) $ is uniformly bounded, or $ \cX $ is compact and $ \bx \mapsto C(\bx) $ is continuous. 
	Assumption \ref{asm:diff} holds if $ \bX $ has a diffuse distribution, \textit{i.e.}, $ \bX $ doesn't contain points with positive mass; and $ \bx \mapsto C(\bx) $ is injective. Assumption \ref{asm:diff} is the key assumption to bound $ \lambda $ away from 0. This assumption will not be necessary if $ k = +\infty $ and $ k^{\star} = 1 $. Assumption \ref{asm:cte} can be met if $ \cX $ is compact and $ \widehat{C}_{n} $ is a random forest estimate \citep{wager2015adaptive}. Following \citet[Theorem 2.7]{steinwart2007fast}, Assumption \ref{asm:prox} can be shown valid if the Tsybakov’s noise assumption on the population margin is met and the kernel bandwidth parameter is chosen appropriately. In the following proposition, we establish the regret bound.

	\begin{prop}[Regret Bound] \label{prop:regret} 
		Suppose $ \cR^{k}_{c} $, $ \cR^{k}_{c,\psi} $, $ \cL^{k}_{c} $ and $ \cL^{k}_{c,\psi} $ are defined as above. Fix $ k \in (1,+\infty] $, $ k^{\star} = {k \over k-1} $, $ c > 1 $. Assume that Assumptions \ref{asm:bdd}-\ref{asm:prox} hold. Let 
		\[ (\widehat{f}_{n},\widehat{\eta}_{n},\widehat{\lambda}_{n}) \in \argmin\limits_{f \in \cF,\eta \in \bbR,\lambda \ge 0}\left\{ \bbE_{n}\ell_{c,\psi}^{k}(f,\eta,\lambda;\widehat{C}_{n}): \|f\|_{\cF} \le \gamma_{n} \right\}, \]
		with the tuning parameter $ \gamma_{n} $ satisfying $ \gamma_{n} = \cO(n^{-{1 \over 2\beta+1}}) $ as $ n \to \infty $. Then there exists constants $ K_{0} = K_{0}(c,M) < +\infty $ and $K_{1} = K_{1}(c,M) < +\infty $ such that for $ 0 < \delta < 1 $, with probability at least $ 1 - \delta $,  we have
		\[ \cR^{k}_{c}(\widehat{f}_{n}) - \cR^{k,*}_{c} \le \cL^{k}_{c}(\widehat{f}_{n},\widehat{\eta}_{n},\widehat{\lambda}_{n}) - \cR^{k,*}_{c} \le K_{0}\sqrt{\log(2/\delta)}n^{-{\beta \over 2\beta+1}} + K_{1}\|\widehat{C}_{n} - C\|_{\infty}. \]
		In particular, there exists $ K_{01},K_{02},K_{11},K_{12} < +\infty $ not depending on $ c,M $, such that
		\[ K_{0}(c,M) = \begin{cases}
		K_{01}{c^{{(k^{\star}+1)(2k^{\star}-1) \over k^{\star}-1} + {1 \over 2}} \over (c-1)^{k^{\star} + 1/2}}M^{k^{\star} + 1/2}, & k < +\infty;\\
		K_{02}cM^{3/2}, & k = +\infty;
		\end{cases} \quad K_{1}(c,M) = \begin{cases}
		K_{11}{c^{2k^{\star}+1} \over (c-1)^{k^{\star}-1}}M^{k^{\star}-1}, & k <+\infty;\\
		K_{12}c, & k = +\infty.
		\end{cases} \]
	\end{prop}

	In Proposition \ref{prop:regret}, it can be of theoretical interest to understand how the regret bound depends on the DR-constant $ c $ and the power order $ k $. Specifically, 
	as $ c \to +\infty $, $ \eta $ approaches to the essential supremum of $ [C(\bX) - \eta]_{+}^{k^{\star}}{\psi[f(\bX)] \over 2} + [-C(\bX) - \eta]_{+}^{k^{\star}}{\psi[-f(\bX)] \over 2} $ \citep[Example 2.3]{krokhmal2007higher}. Then $ \lambda $ vanishes to 0 so that $ 1/\lambda $ tends to $ +\infty $. Since the Lipschitz constant of $ \ell^{k}_{c,\psi}(f,\eta,\lambda) $ with respect to $ \lambda $ scales with $ 1/\lambda^{k^{\star}} $, the universal constants $ K_{0} $ and $ K_{1} $ grow to $ +\infty $ as well. 
	
	Another important fact is that the conjugate number $ k^{\star} $ of $ k $ appears in the polynomial orders of $ c $ and $ M $ respectively in the universal constants $ K_{0} $ and $ K_{1} $. In particular, for a large conjugate order $ k^{\star} $, the universal constants $ K_{0} $ and $ K_{1} $ increase with the DR-constant $ c $ and the CTE bound $ M $ more rapidly. In order to achieve a tighter finite sample regret bound, a smaller $ k^{\star} $ and hence a larger $ k $ is preferred. Such a phenomenon complements the fact that the power uncertainty set $ \cP^{k}_{c}(\bbP) $ decreases in $ k $. Specifically, as the power order $ k $ increases, its conjugate order $ k^{\star} $ decreases, and the regret bound in Proposition \ref{prop:regret} becomes tighter. On the contrary, the power uncertainty set $ \cP^{k}_{c}(\bbP) $ gets smaller, and the worst-case objective is less distributionally robust. Therefore, the power order $ k $ trades off between the distributional robustness in terms of the size of $ \cP^{k}_{c}(\bbP) $, and the finite sample regret bound.
		
	\section{Simulation Studies}\label{sec:simulation}
	In this section, we carry out two simulation studies to evaluate the generalizability of the DR-ITR on the testing distributions that are different from the training distribution. The first simulation considers the covariat shifts. The second simulation considers the mixture of subgroups. 
	
	\subsection{Covariate Shifts}\label{sec:shift}
	In this section, we extend the motivating example in Section \ref{sec:eg} to a more practical simulation setting. Consider the training data generating process: $ n=1,000 $, $ p = 10 $, $ \bX\sim \cN_{p}(\bzero_{p},\Ib_{p}) $, $ A|\bX \sim \Bern(1/2) $ and $ Y|(\bX,A) = m(\bX) + (A-1/2)C(\bX) + \cN(0,1) $, where $ m(\bx) = 1+{1 \over p}\sum_{j=1}^{p}x_{j} $, $ C(\bx) = x_{2} - (x_{1}^{3} - 2x_{1}) $. 
	
	At the training stage, we first obtain a CTE function estimate $ \widehat{C}_{n} $ by fitting a casual forest \citep{wager2018estimation} on the training data. Then we obtain the out-of-bag prediction at the training covariates $ \widehat{C}_{n}(\bX) $. Next we fit the standard ITR by empirically minimizing $ \bbE_{n}\big\{\widehat{C}_{n}(\bX)\left( \psi[f(\bX)]-1\right)\big\} $ as the $ \psi $-relaxation of the empirical risk function $ \bbE_{n}\big\{\widehat{C}_{n}(\bX)\bsign[-f(\bX)]\big\} $, over the linear function class $ \cF_{\gamma}:=\{ f(\bx)=b + \bbeta^{\intercal}\bx: b \in \bbR,\ \bbeta \in \bbR^{p},\ \|\bbeta\|_{2} \le \gamma \} $. The tuning parameter $ \gamma \ge 0 $ is determined by 10-fold cross-validation among $ \{ 0.1,0.5,1,2,4 \} $. Finally, we fit the DR-ITRs for $ k = 2 $ and $ c \in \cC = \{ 1.19,1.38,\cdots,20 \} $ from the function class $ \cF_{\gamma} $, where $ \gamma $ is the same as that of the standard ITR. 
	
	We consider the mean-shifted testing distribution $ \bX \sim \cN_{p}(\bmu,\Ib_{p}) $ for various covariate centroids $ \bmu  $'s. In order to calibrate the DR-constant $ c $ for every fixed $ \bmu $, we generate a calibrating dataset of size $ n_{\rm calib} = 50 $ from the testing distribution. The following two scenarios for the calibrating data are considered here: 1) a randomized controlled trial (RCT) dataset $ (\bX,A,Y) $ is generated, with $ \bX \sim \cN_{p}(\bmu,\Ib_{p}) $ and $ (A,Y) $ as before; and 2) only the covariate vector $ \bX \sim \cN_{p}(\bmu,\Ib_{p}) $ is generated. In Scenario 1, we use the IPWE of the calibrating value function $ \widehat{\cV}_{\rm calib}^{\IPWE}(\widehat{f}_{c}) := \bbE_{n_{\rm calib}}\{Y\bbone[(2A-1)\widehat{f}_{c}(\bX) > 0]/(1/2)\} $ to evaulate the DR-constant $ c $, while in Scenario 2, we use the CTE-based calibrating value function $ \widehat{\cV}^{\CTE}_{\rm calib}(\widehat{f}_{c}) := \bbE_{n_{\rm calib}}\{\widehat{C}_{n}(\bX)\bsign[\widehat{f}_{c}(\bX)]\} $ instead. Here, the estimated CTE function $ \widehat{C}_{n} $ is obtained from the training stage.
	
	For comparison, we consider the following: 1) the LB-ITR that maximizes the value function under the testing distribution; 2) the $ \ell_{1} $-penalized least-square ($ \ell_{1} $-PLS) \citep{qian2011performance} of $ Q(\bX,A) = \bbE(Y|\bX,A) $ on $ (1,\bX,A,A\bX) $ and the corresponding estimated ITR $ \widehat{d}(\bx) \in \argmin_{a \in \{ \pm1 \}}\widehat{Q}_{n}(\bx,a) $; 3) the standard ITR; 4) the RCT-DR-ITR for the calibrating Scenario 1; and 5) the CTE-DR-ITR for the calibrating Scenario 2. We compare the testing values $ \bbE_{n_{\test}}[C(\bX)\widehat{d}(\bX)] $ based on an independent testing dataset of size $ n_{\test}=100,000 $ for every testing distribution. The testing values across different testing distributions are not comparable. For a specific testing distribution, the LB-ITR can be a benchmark to be compared to, since its testing value is the best achievable in theory among the linear ITR class. The training-calibrating-testing procedure is replicated for 500 times. The testing values (standard errors) for $ n_{\rm calib} = 50 $ are reported in Table \ref{tab:shift50}. 
	
	When the testing distribution is the same as training $ (\mu_{1},\mu_{2}) = (0,0) $, the calibration procedures for the DR-ITRs are expected to choose $ c = 1 $, which corresponds to the standard ITR. With the finite calibrating sample, some DR-constant $ c $ greater than 1 can be possibly chosen, leading to smaller testing values for the DR-ITRs in Table \ref{tab:shift50}. In particular, the testing value of the CTE-DR-ITR is higher than that of the RCT-DR-ITR, and is closer to the testing value of the standard ITR in this case. The reason is that, the RCT-based calibrating value function estimate $ \widehat{\cV}^{\rm IPWE}_{\rm calib} $ depends on $ (\bX,A,Y) $ in the calibrating data, while the CTE-based one $ \widehat{\cV}^{\rm CTE}_{\rm calib} $ depends on $ \bX $ only. As a consequence, the CTE-based calibration can be more accurate than the RCT-based one. 
	
\begin{table}[H]
	
	\caption{\label{tab:shift50}Testing Values (Standard Errors) on the Mean-Shifted Covariate Domains ($n_{\rm calib} = 50$)}
	\centering
	\fontsize{7}{5}\selectfont
	\begin{threeparttable}
		\begin{tabular}[t]{>{\raggedleft\arraybackslash}p{1cm}|>{\raggedright\arraybackslash}p{2cm}|>{\raggedright\arraybackslash}p{2.5cm}|>{\raggedright\arraybackslash}p{2.5cm}|>{\raggedright\arraybackslash}p{2.5cm}|>{\raggedright\arraybackslash}p{2.5cm}}
			\toprule
			\diagbox[width=1.25cm,height=0.75cm]{$ \mu_{2} $}{$\mu_{1} $} & type & 0 & 0.734 & 1.469 & 1.958\\
			\rowcolor{gray!6}
			\midrule
			& LB-ITR & \em{2.333} (0.00244) & \em{2.907} (0.011) & \em{5.334} (0.0362) & \em{9.27} (0.0154)\\
			
			& $\ell_{1}$-PLS & \textbf{2.124} (0.0022) & 2.235 (0.011) & 3.613 (0.0505) & 6.32 (0.103)\\
			
			& Standard ITR & 2.089 (0.00158) & 1.735 (0.013) & 1.348 (0.0595) & 1.567 (0.13)\\
			
			& RCT-DR-ITR & 2.085 (0.00444) & 2.286 (0.0114) & 4.545 (0.0255) & 8.371 (0.0451)\\
			
			\multirow{-5}{1cm}{\raggedleft\arraybackslash 1.958} & CTE-DR-ITR & 2.098 (0.00348) & \textbf{2.304} (0.0106) & \textbf{4.551} (0.0238) & \textbf{8.459} (0.0424)\\
			\cmidrule{1-6}
			\rowcolor{gray!6}
			& LB-ITR & \em{1.893} (0.00712) & \em{2.627} (0.00656) & \em{5.28} (0.0213) & \em{9.379} (0.0128)\\
			
			& $\ell_{1}$-PLS & 1.667 (0.00307) & \textbf{2.021} (0.0076) & 4.095 (0.0342) & 7.573 (0.0706)\\
			
			& Standard ITR & \textbf{1.674} (0.00152) & 1.645 (0.0127) & 2.377 (0.0553) & 4.011 (0.119)\\
			
			& RCT-DR-ITR & 1.627 (0.00688) & 1.987 (0.00997) & 4.484 (0.0192) & 8.611 (0.0285)\\
			
			\multirow{-5}{1cm}{\raggedleft\arraybackslash 1.469} & CTE-DR-ITR & 1.663 (0.00326) & 1.997 (0.00992) & \textbf{4.55} (0.0163) & \textbf{8.686} (0.0269)\\
			\cmidrule{1-6}
			\rowcolor{gray!6}
			& LB-ITR & \em{1.227} (0.00244) & \em{2.144} (0.00609) & \em{5.269} (0.00931) & \em{9.608} (0.00898)\\
			
			& $\ell_{1}$-PLS & 1.094 (0.00418) & \textbf{1.676} (0.00442) & 4.587 (0.0151) & 8.8 (0.0314)\\
			
			& Standard ITR & \textbf{1.174} (0.00149) & 1.553 (0.00806) & 3.739 (0.0379) & 7.06 (0.0763)\\
			
			& RCT-DR-ITR & 1.094 (0.00753) & 1.651 (0.00675) & 4.622 (0.0109) & 9.036 (0.015)\\
			
			\multirow{-5}{1cm}{\raggedleft\arraybackslash 0.734} & CTE-DR-ITR & 1.152 (0.00292) & 1.667 (0.00588) & \textbf{4.648} (0.0113) & \textbf{9.06} (0.0161)\\
			\cmidrule{1-6}
			\rowcolor{gray!6}
			& LB-ITR & \em{0.9942} (0.00202) & \em{1.774} (0.0034) & \em{5.232} (0.00559) & \em{9.767} (0.0068)\\
			
			& $\ell_{1}$-PLS & 0.8296 (0.00454) & 1.648 (0.0036) & \textbf{4.914} (0.00501) & \textbf{9.476} (0.0103)\\
			
			& Standard ITR & \textbf{0.9437} (0.00153) & 1.679 (0.00336) & 4.654 (0.017) & 8.895 (0.0342)\\
			
			& RCT-DR-ITR & 0.8374 (0.00821) & 1.647 (0.00574) & 4.868 (0.00797) & 9.444 (0.00841)\\
			
			\multirow{-5}{1cm}{\raggedleft\arraybackslash 0.000} & CTE-DR-ITR & 0.9206 (0.00272) & \textbf{1.688} (0.00289) & 4.888 (0.00698) & 9.442 (0.00999)\\
			\bottomrule
		\end{tabular}
		\begin{tablenotes}
			\item[1] $\bmu = (\mu_{1},\mu_{2},0,\cdots,0)^{\intercal}$ with $ \mu_{1} $ in column and $ \mu_{2} $ in row is the testing covariate centroid.
			\item[2] Values (larger the better) can be comparable for the same $ (\mu_{1},\mu_{2}) $ but incomparable across different $ (\mu_{1},\mu_{2}) $.
			\item[3] LB-ITR maximizes the testing value function at $ (\mu_{1},\mu_{2}) $ over the linear ITR class. The corresponding testing value is the best achievable among the linear ITR class.
		\end{tablenotes}
	\end{threeparttable}
\end{table}

\begin{figure}[!h]
		\centering
		\begin{subfigure}{0.49\linewidth}
			\includegraphics[width=\linewidth]{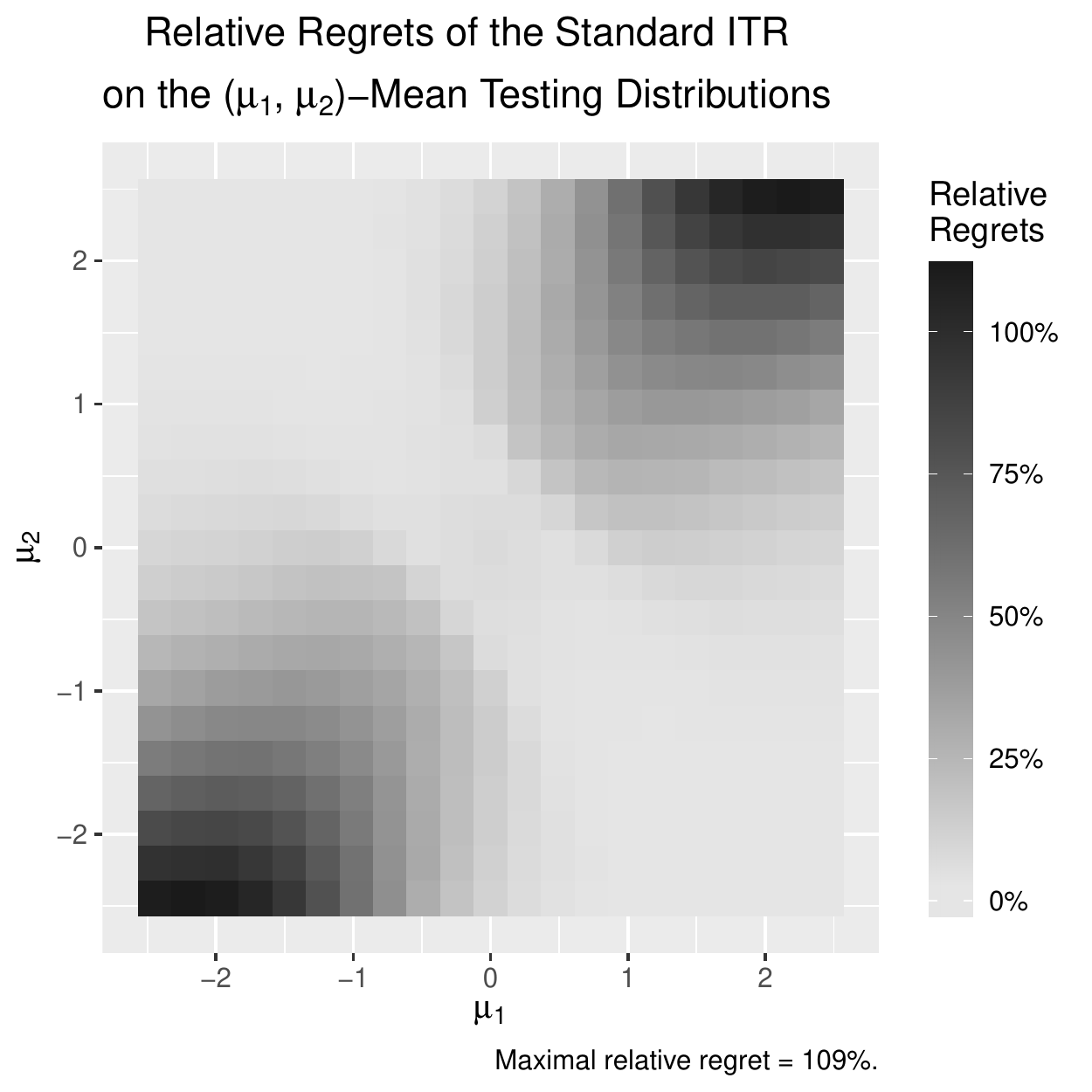}
			\caption{Standard ITR}
			\label{fig:nonlin_test_regret_ERM}
		\end{subfigure}
		\begin{subfigure}{0.49\linewidth}
			\includegraphics[width=\linewidth]{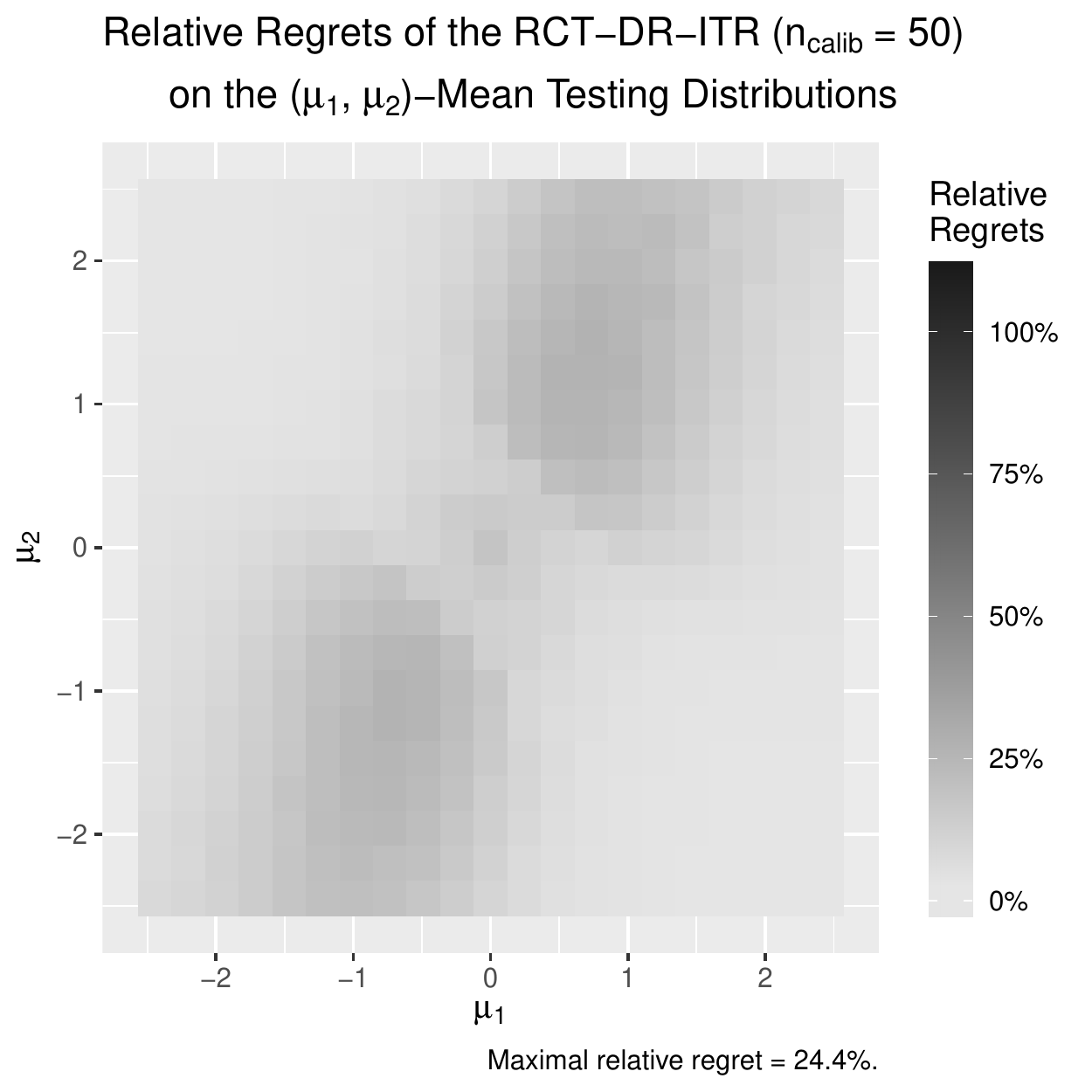}
			\caption{RCT-DR-ITR ($ n_{\rm calib} = 50 $)}
			\label{fig:nonlin_test_regret_calib_outcome50}
		\end{subfigure}
		\caption{Relative Regrets on the Mean-Shifted Covariate Domains (lighter the better).}
		\label{fig:nonlin_test_regret}
	\end{figure}

\begin{figure}[!h]
		\centering
		\includegraphics[width=0.8\linewidth]{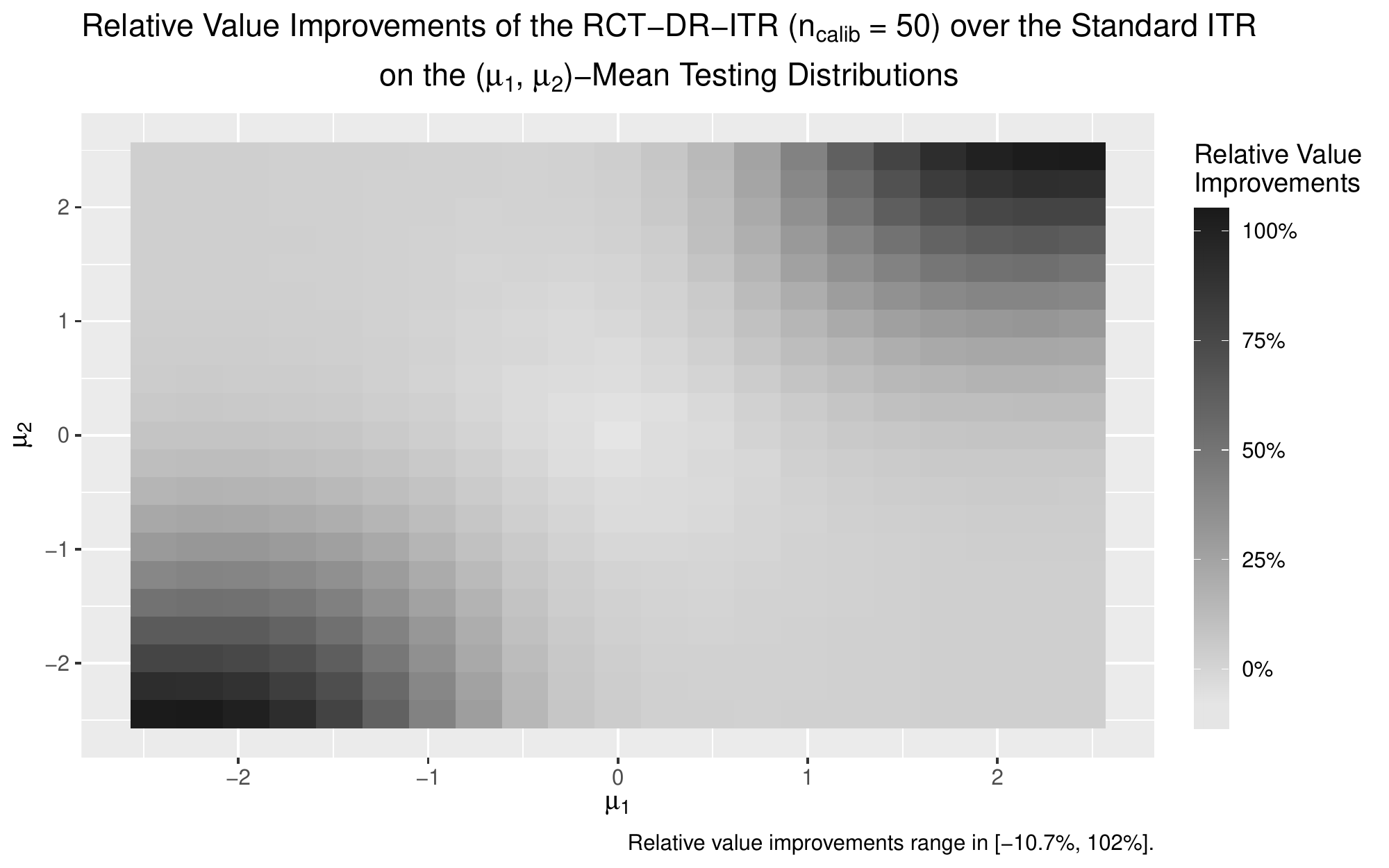}
		\caption{Relative improvements of the RCT-DR-ITR over the standard ITR as the difference of their relative regrets on the mean-shifted covariate domains ($ n_{\rm calib} = 50 $, darker the better).}
		\label{fig:nonlin_test_value_impv_calib_outcome50}
	\end{figure}

	When $ (\mu_{1},\mu_{2}) \ne (0,0)$, the testing distribution is different from training, and the performance of the standard ITR deteriorates while the DR-ITRs still maintain reasonably good performance. The phenomenon is more evident when $ \mu_{1},\mu_{2} \in \{1.469, 1.958\} $. In particular at $ (\mu_{1},\mu_{2}) = (1.958,1.958) $, the value of the standard ITR can be as low as 17\% of the best achievable value among the linear ITR class, while the DR-ITRs can maintain more than 90\%. In fact, such a phenomenon is general. In Figure \ref{fig:nonlin_test_regret_ERM}, we further enumerate the testing covariate centroid $ \bmu = (\mu_{1},\mu_{2},0,\cdots,0)^{\intercal} $ for $ \mu_{1},\mu_{2} \in [-2.448,2.448] $ and compute the relative regrets of the standard ITR and the RCT-DR-ITR. Across all mean-shifted testing distributions, the relative regrets of the standard ITRs can be as high as 108\%, in which case the standard ITR value is negative, and hence even worse than the completely random treatment rule $ d_{\rand} $. On the contrary, the relative regrets for the RCT-DR-ITR ($ n_{\rm calib} = 50$) shown in Figure \ref{fig:nonlin_test_regret_calib_outcome50} are at most 24\% across all testing centroids. 
	This suggests that the RCT-DR-ITR maintains relatively good performance on all such testing distributions, while the standard ITR fails. Figure \ref{fig:nonlin_test_value_impv_calib_outcome50} further shows that the DR-ITR provides substantial testing value improvements over the standard ITR. This demonstrates that the small sample size $ n_{\rm calib} = 50 $ is sufficient for calibrating the DR-ITR with significant testing improvement. 
	
	From Table \ref{tab:shift50}, it can be also observed that $ \ell_{1} $-PLS can have better performance than the standard ITR when training and testing distributions are different. The reason is that, the objective of $ \ell_{1} $-PLS does not target the value function under the training distribution directly, but rather, the mean squared error of the linear approximation to $ Q(\bX,A) $ under the training distribution. Such a linear approximation can perform well when the testing distribution is not far from the training distribution. However, in the case $ \mu_{1},\mu_{2} \in \{1.469, 1.958\} $ in the sense that the testing distribution deviates more from the training one, the DR-ITRs enjoy notably higher testing values than $ \ell_{1} $-PLS.
	
	In the Supplementary Material, we provide more detailed results for other comparisons including the relative regrets/improvements on all mean-shifted covariate domains of all centroids, the misclassification rates on all mean-shifted covariate domains of all centroids, the comparison with some other methods in relative regrets and misclassification rates, and the case of $ k \in \{ 1.25, 1.5, 2, 3, \infty \} $.
	In particular, the misclassification rates inform similar conclusions as the relative regrets/improvements. If we increase the calibrating sample size from 50 to 100, then the testing values of DR-ITRs can be further improved. We also find that among our simulation scenarios, the testing values of the DR-ITR are not very sensitive to difference choices of $ k $.
	
	\subsection{Performance on the Mixture of Subgroups}\label{sec:mix}
	In this section, we consider a population that consists of two subgroups, with each following a distinct CTE function. We aim to find an ITR that can generalize well on different mixtures of subgroups.
 	
	We modify the simulation setup in Section \ref{sec:shift} as follows: $ \bX|\xi \sim \xi\cN_{p}(\bmu_{1},\Ib_{p}) + (1-\xi)\cN_{p}(\bmu_{0},\Ib_{p}) $, where $ \xi \sim \Bern(p_{\mix}) $ is the unobservable mixture/subgroup indicator with subgroup 1 probability $ p_{\mix} $ and subgroup 0 probability $ 1-p_{\mix} $, and the subgroup means $ \bmu_{1} = (-1/2,1/2,0,\cdots,0)^{\intercal} $ and $ \bmu_{0} = -\bmu_{1} $. We consider the CTE function $ C(\bx;\xi) := (2\xi - 1)\beta_{0} + \beta_{1}x_{1} + \beta_{2}x_{2} $ that is linear in the covariate vector, but with a subgroup-dependent intercept $ (2\xi - 1)\beta_{0} $, and $ (\beta_{0},\beta_{1},\beta_{2}) := (-3/2,-2,1) $. The unconditional CTE function is nonlinear:
	\[ C(\bx) := \bbE[C(\bx;\xi)|\bX = \bx] = {p_{\mix}\exp(-\| \bx - \bmu_{1} \|_{2}^{2}/2) - (1-p_{\mix})\exp(-\| \bx - \bmu_{0} \|_{2}^{2}/2) \over p_{\mix}\exp(-\| \bx - \bmu_{1} \|_{2}^{2}/2) + (1-p_{\mix})\exp(-\| \bx - \bmu_{0} \|_{2}^{2}/2)}\beta_{0} + \beta_{1}x_{1} + \beta_{2}x_{2}. \]
	In particular, the unconditional CTE function $ C(\bx) $ depends on the subgroup 1 probability $ p_{\mix} $. The distributional changes are due to the subgroup 1 probability. Specifically, the training subgroup 1 probability is $ 0.75 $, while the testing subgroup 1 probability varies in $ \{ 0.1,0.25,0.5,0.75,0.9 \} $. Since the training and testing CTE functions can be different, Assumption \ref{asm:cov} cannot be fully met. Therefore, our proposed DR-ITR can be robust to such distributional changes only to some extent.
	
	We consider the same training-calibrating-testing procedure as that in Section \ref{sec:shift}, except that the DR-constant $ c $ ranges in $ \{ 1.18,1.27,\cdots,10 \} $. The testing values of the ITRs are reported in Table \ref{tab:mix50}. When the training and testing distributions are the same at $ p_{\mix} = 0.75 $, all ITRs have similar testing performance. The standard ITRs have higher testing values than the DR-ITRs in this case. When the testing $ p_{\mix} $ becomes smaller, the DR-ITRs show better testing performance than the standard ITR. When the testing $ p_{\mix} = 0.25 $ or $ 0.1 $, the RCT-DR-ITR has the highest testing values among all. Since the true testing CTE function changes along with the testing $ p_{\mix} $, the corresponding estimate $ \widehat{C}_{n} $ based on the training data can suffer from the generalizability problem. Therefore, the CTE-based calibration performs slightly worse than the RCT-based calibration in this case. However, the CTE-based DR-ITR is superior to the standard ITR, and is comparable to the $ \ell_{1} $-PLS. More detailed comparisons and the case $ n_{\rm calib} = 100 $ are provided in the Supplementary Material.
	
\begin{table}[!h]
	
	\caption{\label{tab:mix50}Testing Values (Standard Errors) on the Mixture of Subgroups ($n_{\rm calib} = 50$)}
	\centering
	\fontsize{7}{7}\selectfont
	\begin{threeparttable}
		\begin{tabular}[t]{>{\raggedright\arraybackslash}p{2cm}|>{\raggedright\arraybackslash}p{2.25cm}|>{\raggedright\arraybackslash}p{2.25cm}|>{\raggedright\arraybackslash}p{2.25cm}|>{\raggedright\arraybackslash}p{2.25cm}|>{\raggedright\arraybackslash}p{2.25cm}}
			\toprule
			\multicolumn{1}{c}{\textbf{ }} & \multicolumn{5}{c}{\textbf{Testing Subgroup 1 Probability}} \\
			\cmidrule(l{3pt}r{3pt}){2-6}
			\textbf{type} & \textbf{0.1} & \textbf{0.25} & \textbf{0.5} & \textbf{0.75} & \textbf{0.9}\\
			\rowcolor{gray!6}
			\midrule
			LB-ITR & \em{1.665} (0.0067) & \em{1.537} (0.00618) & \em{1.444} (0.00412) & \em{1.545} (0.00537) & \em{1.679} (0.00585)\\
			$\ell_{1}$-PLS & 1.182 (0.00191) & 1.264 (0.0014) & \textbf{1.399} (0.000591) & \textbf{1.537} (0.000333) & 1.624 (0.000781)\\
			Standard ITR & 1.143 (0.00434) & 1.232 (0.00329) & 1.383 (0.0015) & 1.535 (0.000543) & \textbf{1.632} (0.00142)\\
			RCT-DR-ITR & \textbf{1.267} (0.0066) & \textbf{1.305} (0.00423) & 1.395 (0.00256) & 1.52 (0.00212) & 1.614 (0.00234)\\
			CTE-DR-ITR & 1.16 (0.00409) & 1.247 (0.00323) & 1.388 (0.00137) & 1.534 (0.00055) & 1.628 (0.00149)\\
			\bottomrule
		\end{tabular}
		\begin{tablenotes}
			\item[1] Testing subgroup 1 probability = 0.75 is the same as the training one.
			\item[2] Values (larger the better) can be comparable for the same subgroup 1 probability but incomparable across different subgroup 1 probabilities
			\item[3] LB-ITR maximizes the testing value function over the linear ITR class. The corresponding testing value is the best achievable among the linear ITR class.
		\end{tablenotes}
	\end{threeparttable}
\end{table}
	\section{Application to the ACTG 175 Trial Data}\label{sec:aids}
	In this section, we evaluate the generalizability of our proposed DR-ITR on a clinical trial dataset from the ``AIDS clinical trial group study 175" \citep{hammer1996trial}. The goal of this study was to compare four treatment arms among 2,139 randomly assigned subjects with human immunodeficiency virus type 1 (HIV-1), whose CD4 counts were 200-500 cells/mm$ ^{3} $. The four treatments are the zidovudine (ZDV) monotherapy, the didanosine (ddI) monotherapy, the ZDV combined with ddI, and the ZDV combined with zalcitabine (ZAL). 
	
	The evidence found from the AIDS trial data can have some generalizability problems. When studying women living with HIV and women at risk for HIV infection in the USA cohort, the Women’s Interagency HIV Study (WIHS) \citep{bacon2005women} has been considered to be representative. However, it was reported in \citet{gandhi2005eligibility} that 28-68\% of the HIV positive women in WIHS were excluded from the eligibility criteria of many ACTG studies. In the ACTG 175 dataset, the number of female patients is only 368 out of 2139. Thus we suspect that the female patients may be underrepresented in this dataset, and the ITR based on the dataset may not generalize well on the women subgroup. In this section, we study the generalizability of DR-ITR when the testing dataset consists of female patients only. Specifically, the training dataset is a subsample from ACTG 175 with original male/female proportion, while the testing dataset is a subsample from the female patients of ACTG 175, and there is no overlap across training and testing. We try to resemble the ideal world that we can have independent testing data from the female population.
	
	We consider the outcome $ Y $ as the difference between the early stage (at 20$ \pm $5 weeks from baseline) CD4 cell counts and the CD4 counts at baseline. We focus on the treatment comparison between the ZDV + ZAL ($ A = 1 $) and the ddI ($ A = -1 $), and the corresponding patients from the dataset. In particular, only 180 of them are women. The average treatment effects on the male and female subgroups are $ -8.97 $ and $ -1.39 $ respectively, which suggests that there is treatment effect discrepancy between these subgroups. We sample the training data from the ACTG 175 dataset in the ZDV + ZAL or ddI arm of sample size $ 1,085 \times 60\% = 651 $  stratified to the gender. In particular, the training dataset includes $ 180\times 60\% = 108 $ female patients. The remaining female data $ (180-108 = 72) $ are used for testing. We only consider female patients in testing. We further sample 50 from the testing female data for calibration, and the remaining $ (72-50=22) $ are the testing dataset. We also consider 12 selected baseline covariates $ \bX $ as was studied in \citet{lu2013variable}. There are 5 continuous covariates: age (year), weight (kg, coded as \texttt{wtkg}), CD4 count (cells/mm$ ^{3} $) at baseline, Karnofsky score (scale of 0-100, coded as \texttt{karnof}), CD8 count (cells/mm$ ^{3} $) at baseline. They are centered and scaled before further analysis. In addition, there are 7 binary variables: gender (1 = male, 0 = female), homosexual activity (\texttt{homo}, 1 = yes, 0 = no), race (1 = nonwhite, 0 = white), history of intravenous drug use (\texttt{drug}, 1 = yes, 0 = no), symptomatic status (\texttt{symptom}, 1 = symptomatic, 0 = asymptomatic), antiretroviral history (\texttt{str2}, 1 = experienced, 0 = naive) and hemophilia (\texttt{hemo}, 1 = yes, 0 = no).
	
	Before fitting ITRs, we estimate the CTE function $ C(\bX) $ by the following regress-and-subtract procedure: first we fit two separate random forests by regressing $ Y $ on $ \bX $ restricted on $ A = 1 $ and $ A = -1 $ respectively; then we subtract two regression models to obtain the CTE function estimate $ \widehat{C}_{n}(\bX) $. We follow the same implementation as in Section \ref{sec:shift} to fit the standard ITR and DR-ITRs over a constrained linear function class $ \cF_{\gamma}:=\{ f(\bx)=b + \bbeta^{\intercal}\bx: b \in \bbR,\ \bbeta \in \bbR^{p},\ \|\bbeta\|_{2} \le \gamma \} $ on the training data. The testing performance is evaluated by the IPWE of the value function on the testing data. The training-calibrating-testing procedure is repeated for 1,500 times. The testing values are reported in Table \ref{tab:ACTG175_test}, where the value can be interpreted as the expected CD4 count improvement from baseline at the early stage ($ 20\pm 5 $ weeks). In addition to the calibrated DR-ITRs, we also include the value of the \textit{best DR-ITR} that enjoys the highest testing performance among all DR-constants. For comparison, we include the results of residual weighted learning (RWL) \citep{zhou2017residual} with linear kernel. Both RWL and the standard ITR share similar implementation, except that RWL can be shown equivalently using $ \widehat{C}_{n}(\bX) = \widehat{Q}_{n}(\bX,1) - \widehat{Q}_{n}(\bX,-1) + 2A[Y-\widehat{Q}_{n}(\bX,A)] $ as a plug-in CTE estimate.
	
\begin{figure}[!h]
		\centering
		\includegraphics[width=0.8\linewidth]{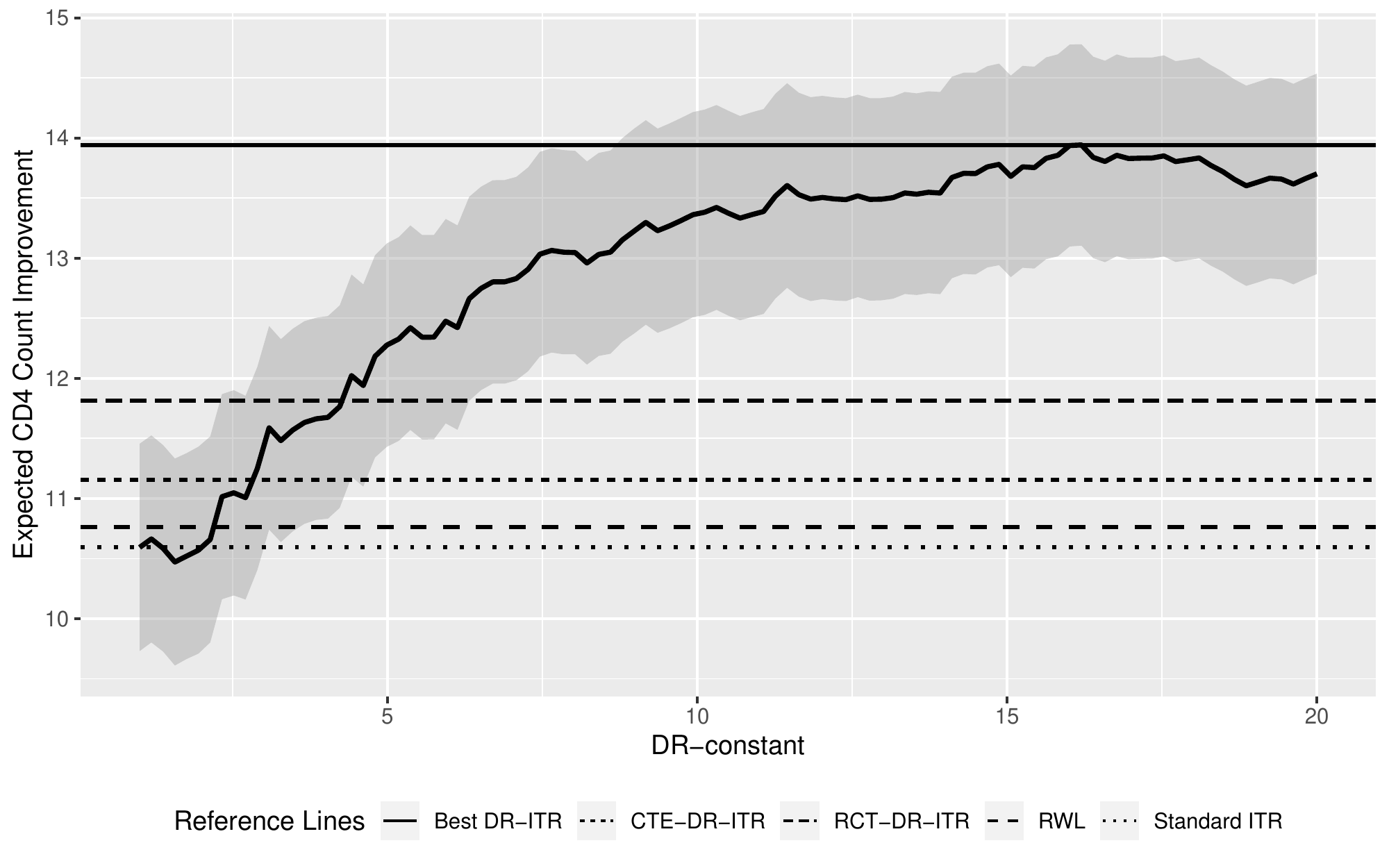}
		\caption{Expected CD4 Count Improvement (cells/mm$ ^{3} $) from Baseline at the Early Stage (20$ \pm $5 weeks) of the DR-ITRs of Various DR-Constants on the ACTG 175 Female Patients (higher the better)}
		\label{fig:ACTG175_test}
	\end{figure}

	The testing results show that our proposed DR-ITRs can have better values than the standard ITR and RWL. In particular, the improvement of the best DR-ITR is substantial, while the improvements of the calibrated ITRs are not as strong. We plot the testing values of the DR-ITRs against the corresponding DR-constants in Figure \ref{fig:ACTG175_test}. It suggests that the testing values generally increase with the DR-constant. In this analysis, the calibrated DR-constants are not close to the optimal DR-constant. As a result, the testing performance of the calibrated DR-ITRs is not as good as the best DR-ITR. One reason for this phenomenon can be that the outcome $ Y $ has a heavy tail distribution, as was highlighted in \cite{qi2019estimating}, so that the value function estimate is highly variable based on the small calibrating sample. Another reason can be that the random forest regress-and-subtract estimate of the CTE function does not generalize well on the testing distribution.

\begin{table}[!h]

\caption{\label{tab:ACTG175_test}Expected CD4 Count Improvement (cells/mm$ ^{3} $) from Baseline at the Early Stage (20$ \pm $5 weeks) and Standard Errors on the ACTG-175 Female Patients (higher the better).}
\centering
\fontsize{8}{10}\selectfont
\begin{tabu} to \linewidth {>{\centering}X>{\centering}X>{\centering}X>{\centering}X>{\centering}X}
\toprule
RWL & Standard ITR & Best DR-ITR & RCT-DR-ITR & CTE-DR-ITR\\
\midrule
10.7617 (0.8636) & 10.593 (0.8627) & \textbf{13.9423} (0.8378) & 11.8133 (0.8357) & 11.1563 (0.8514)\\
\bottomrule
\end{tabu}
\begin{itemize}
	\item[] Standard errors are computed based on 1,500 replications.
\end{itemize}
\end{table}
	
	On the overall dataset, we fit the DR-ITRs and report their fitted coefficients in Table \ref{tab:ACTG175_coef} for selected DR-constants. To stabilize the randomness from the random forest estimate of the CTE function, we refit the random forest 20 times and average the corresponding DR-ITR coefficients. We find that there are noticeable changes in the coefficients of the intercept and the homosexual activity when the DR-constant gets large. Within the ACTG 175 dataset (ZDV + ZAL or ddI), we find that only 6 female patients have homosexual activity. Four of them are treated with ZDV + ZAL, and the change of their CD4 counts are 123, 34, $ -11 $ and 158 respectively. Two of them are treated with ddI, and the change of their CD4 counts are $ -41 $, $ -182 $. Therefore, the ZDV + ZAL ($ A = +1 $) may have more benefits compared to the ddI $ (A = -1) $ on these patients. This helps to explain why the larger coefficients in homosexual activity for the larger DR-constants can be beneficial for the female patients.
	
\begin{table}[!h]

\caption{\label{tab:ACTG175_coef}Linear Coefficients of the DR-ITRs Fitted on the ACTG 175 Dataset}
\centering
\fontsize{6}{5}\selectfont
\begin{tabu} to \linewidth {>{\raggedleft}X>{\raggedleft}X>{\raggedleft}X>{\raggedleft}X>{\raggedleft}X>{\raggedleft}X>{\raggedleft}X>{\raggedleft}X>{\raggedleft}X>{\raggedleft}X>{\raggedleft}X>{\raggedleft}X>{\raggedleft}X>{\raggedleft}X}
\toprule
\textbf{DR-constant} & \textbf{Intercept} & \textbf{age} & \textbf{wtkg} & \textbf{cd40} & \textbf{karnof} & \textbf{cd80} & \textbf{gender} & \textbf{homo} & \textbf{race} & \textbf{drugs} & \textbf{symptom} & \textbf{str2} & \textbf{hemo}\\
\midrule
\rowcolor{gray!6}
$1$ & $-0.02$ & $-0.25$ & $0.06$ & $-0.58$ & $-0.06$ & $0.53$ & $-0.16$ & $-0.4$ & $0.16$ & $0.16$ & $0.16$ & $0.16$ & $0.09$\\
$4.8$ & $-0.31$ & $-0.23$ & $0.12$ & $-0.67$ & $0.11$ & $0.55$ & $-0.12$ & $-0.21$ & $0.2$ & $0.12$ & $0.1$ & $-0.06$ & $0.09$\\
$8.6$ & $-0.43$ & $-0.23$ & $0.11$ & $-0.64$ & $0.16$ & $0.54$ & $-0.11$ & $-0.05$ & $0.12$ & $0.04$ & $0.07$ & $-0.24$ & $0.01$\\
$12.4$ & $-0.54$ & $-0.22$ & $0.1$ & $-0.64$ & $0.19$ & $0.51$ & $-0.04$ & $0.01$ & $0.08$ & $0.05$ & $0.04$ & $-0.27$ & $-0.02$\\
\rowcolor{gray!6}
$16.2$ & $-0.61$ & $-0.23$ & $0.1$ & $-0.64$ & $0.2$ & $0.51$ & $0$ & $0.03$ & $0.06$ & $0.05$ & $0.02$ & $-0.27$ & $-0.02$\\
$20$ & $-0.64$ & $-0.24$ & $0.09$ & $-0.63$ & $0.22$ & $0.5$ & $0.01$ & $0.03$ & $0.05$ & $0.07$ & $0.01$ & $-0.26$ & $-0.01$\\
\bottomrule
\multicolumn{14}{l}{\textsuperscript{1} DR-constant = 1 corresponds to the standard ITR; DR-constant = 16.2 has the highest testing value in Figure \ref{fig:ACTG175_test}.}\\
\end{tabu}
\end{table}
	
	\section{Discussion}\label{sec:discuss}
	
	In this paper, we propose a new framework for learning a distributionally robust ITR by maximizing the worst-case value function among values under distributions within the power uncertainty set. We introduce two possible calibration scenarios under which the DR-constant can be tuned adaptively to a small amount of the calibrating data from the target population. In this way, when the training and testing distributions are identical, the calibrated DR-ITRs can achieve similar performance as compared to the standard ITR. When the testing distribution deviates from the training distribution, we show that there are many possible scenarios that the standard ITR generalizes poorly, while the calibrated DR-ITRs maintain relatively good testing performance. Our simulation studies and an application to the ACTG 175 dataset demonstrate the competitive generalizability of our proposed DR-ITR.
	
	The main assumption on the changes of covariates in our DR-ITR framework is equivalent to the selection unconfoundedness assumption in a randomized controlled trial. In practice, there may exist unmeasured selection confounding problems for the trial data, and the distributional changes affect both the covariates and the CTE function. One possible extension is to consider the simultaneous changes of the covariate distribution and the CTE function, and leverage more general robustness measure against these changes. 
	
	In our DR-ITR framework, we require an estimate of the CTE function based on the flexible nonparametric techniques. The performance of our DR-ITR can depend on the quality of the CTE function estimate. An alternative strategy is to avoid plugging in a CTE estimate. Instead, the dual representation (\ref{eq:dritr_dual}) can be identified from $ (\bX,A,Y) $ directly using a variational representation of $ [\pm C(\bX) - \eta]^{k^{\star}}_{+} $ \citep{duchi2019distributionally}. This can be a possible extension of our framework. 
	
	Another possible extension is to consider the problem of high-dimensional covariates. Our current formulation involves an $ \ell_{2} $-constraint to control the model complexity. It can be extended to obtain sparse solutions when a $ \ell_{1} $-constraint is used instead. Besides the high-dimensional extension, our current theoretical results assume that $ C(\bX) $ is uniformly bounded. It will be interesting to relax the assumption, such as sub-Gaussianity. Further investigations along these lines can be pursued.
	
	\section*{Acknowledgments}
	
	The authors would like to thank the editor, the associate editor, and reviewers, whose helpful comments and suggestions led to a much improved presentation.
	
	\section*{Funding}
	
	The authors were supported in part by NSF grants IIS-1632951, DMS-1821231, and NIH grants R01GM126550 and P01 CA-142538.

	\bibliography{bibfile.bib}
	\bibliographystyle{asa}
\end{document}